\documentclass{article}

\usepackage{arxiv}
\usepackage{color}

\usepackage[utf8]{inputenc} 
\usepackage[T1]{fontenc}    
\usepackage{amsmath} 

\usepackage{hyperref} 
\usepackage{cleveref}
\usepackage{graphicx}
\usepackage{subcaption}
\usepackage{url}            
\usepackage{booktabs}       
\usepackage{amsfonts}       
\usepackage{nicefrac}       
\usepackage{microtype}      
\usepackage{lipsum}
\usepackage{graphicx}
\usepackage{float}

\usepackage{multirow} 
\usepackage{array}    

\usepackage{listings}
\usepackage{xcolor}

\definecolor{codegreen}{rgb}{0,0.6,0}
\definecolor{codegray}{rgb}{0.5,0.5,0.5}
\definecolor{codepurple}{rgb}{0.58,0,0.82}
\definecolor{backcolour}{rgb}{0.95,0.95,0.92}

\lstdefinestyle{mystyle}{
    backgroundcolor=\color{backcolour},   
    commentstyle=\color{codegreen},
    keywordstyle=\color{magenta},
    numberstyle=\tiny\color{codegray},
    stringstyle=\color{codepurple},
    basicstyle=\ttfamily\footnotesize,
    breakatwhitespace=false,         
    breaklines=true,                 
    captionpos=b,                    
    keepspaces=true,                 
    numbers=left,                    
    numbersep=5pt,                  
    showspaces=false,                
    showstringspaces=false,
    showtabs=false,                  
    tabsize=2
}

\graphicspath{ {./images/} }

\newcommand{\R}{\mathbb{R}}
\DeclareMathOperator{\id}{id}
\newcommand{\refe}{\sigma}
\newcommand{\temp}{\mu}

\def\ac{\textnormal{ac}}

\newtheorem{theorem}{Theorem}[section]

\title{Linearized Optimal Transport pyLOT Library: \\ A Toolkit for Machine Learning on Point Clouds}

\author{Jun Linwu\thanks{Equal contribution} \\
  Halıcıo{\u g}lu Data Science Institute \\
  University of California San Diego \\
  La Jolla, CA 92093 \\
\texttt{julinwu@ucsd.edu} \\
  \And
  Varun Khurana$^*$\\
  Department of Applied Mathematics\\
  Brown University \\
  Providence, RI 02912\\
\texttt{varun\_khurana@brown.edu}\\
  \AND
  Nicholas Karris\\
  Department of Mathematics\\
  University of California San Diego \\
  La Jolla, CA 92093 \\ \texttt{nkarris@ucsd.edu} 
  \And
  Alexander Cloninger\\
  Department of Mathematics and \\
  Halıcıo{\u g}lu Data Science Institute \\
  University of California San Diego \\
  La Jolla, CA 92093 \\ \texttt{acloninger@ucsd.edu} 
}

\date{}

\begin{document}
\maketitle

\begin{abstract}
The \textbf{pyLOT} library offers a Python implementation of linearized optimal transport (LOT) techniques and methods to use in downstream tasks.  The pipeline embeds probability distributions into a Hilbert space via the Optimal Transport maps from a fixed reference distribution, and this linearization allows downstream tasks to be completed using off the shelf (linear) machine learning algorithms.  We provide a case study of performing ML on 3D scans of lemur teeth, where the original questions of classification, clustering, dimension reduction, and data generation reduce to simple linear operations performed on the LOT embedded representations.

\end{abstract}


\section{Introduction}

Many modern machine learning techniques are designed to work with data in the form of finite-dimensional vectors \(\{x_i\}_{i=1}^N \subset \mathbb{R}^d\). However, in numerous real-world applications, the data are not best represented by such finite-dimensional vectors. Instead, point clouds or continuous probability measures are the appropriate data structures. These data arise naturally in fields such as computer vision, image processing, shape analysis, and generative modeling, where representing complex objects as probability distributions provides a richer and more flexible framework for analysis. Real-world examples include text documents with bag-of-words models treating word counts as features, which forms a histogram for each document \cite{zhang2010understanding}, imaging data where pixel intensity is interpreted as mass \cite{rubner2000earth} and results in 2D discrete probability measures over the image grid, and gene expression data that is interpretted as a distribution across a gene network \cite{Chen:2017aa,mathews18}.

Optimal transport (OT) theory \cite{villani2008optimal} has recently emerged as a powerful tool to compare probability measures. Qualitatively, OT generates a distance metric between probability measures by minimizing the work needed to move one distribution into another over all transport plans. It has gained significant popularity for applications \cite{arjovsky2017wasserstein,rubner2000earth,solomon2014wasserstein} involving point clouds and probability distributions. OT allows for the computation of distances between distributions by solving a minimization problem over transportation plans. Despite its theoretical elegance and its ability to capture geometric properties of distributions, using vanilla OT is computationally expensive and does not directly integrate into existing machine learning pipelines. For this reason, OT has been somewhat limited in practical applications, particularly in settings that demand scalable and efficient algorithms for tasks such as classification, dimension reduction, and generation.

To overcome these limitations, \emph{Linearized Optimal Transport} (LOT) \cite{wei13,park18, aldroubi20, moosmueller20,gigli-2011, khurana2022supervised} or ``Monge embedding'' \cite{merigot20} provides an effective framework that embeds measure-valued data into a Hilbert space, bridging the gap between optimal transport theory and machine learning so that measure-valued data can be processed using standard, out-of-the-box machine learning algorithms. By ``linearizing'' the problem, LOT offers both computational tractability and the ability to leverage classic techniques in linear spaces, making it highly appealing for practitioners who need tools that are both practical and efficient.

\begin{figure}[ht]
    \centering
    \begin{subfigure}[b]{0.45\linewidth}
        \centering
        \includegraphics[width=\linewidth]{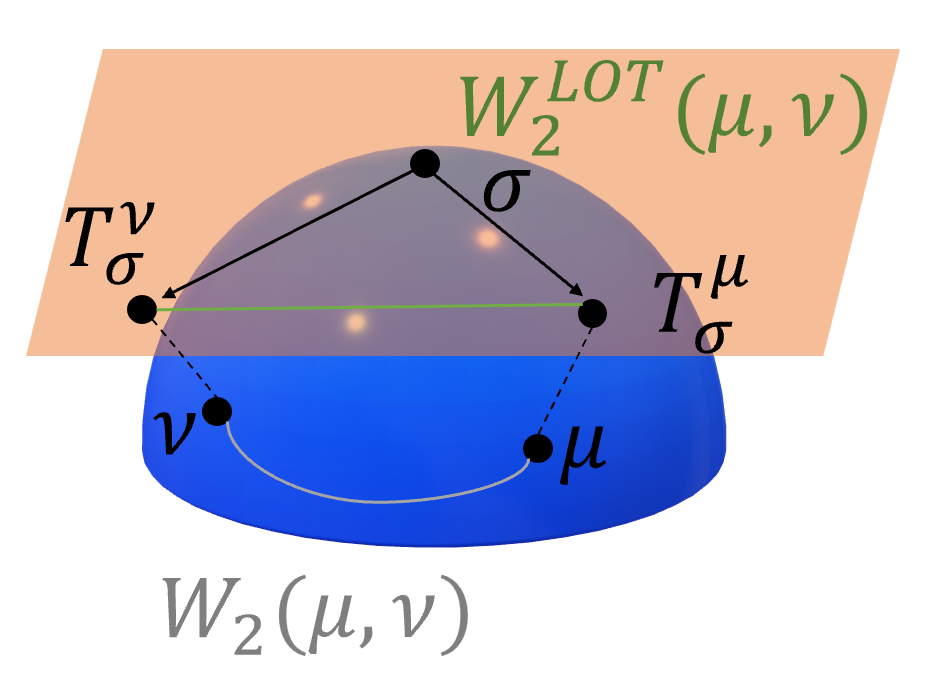}
        \caption{LOT represents inverse exponential map on Wasserstein manifold.}
        \label{fig:lot_manifold}
    \end{subfigure}%
    \hspace{0.05\linewidth} 
    \begin{subfigure}[b]{0.45\linewidth}
        \centering
        \includegraphics[width=\linewidth]{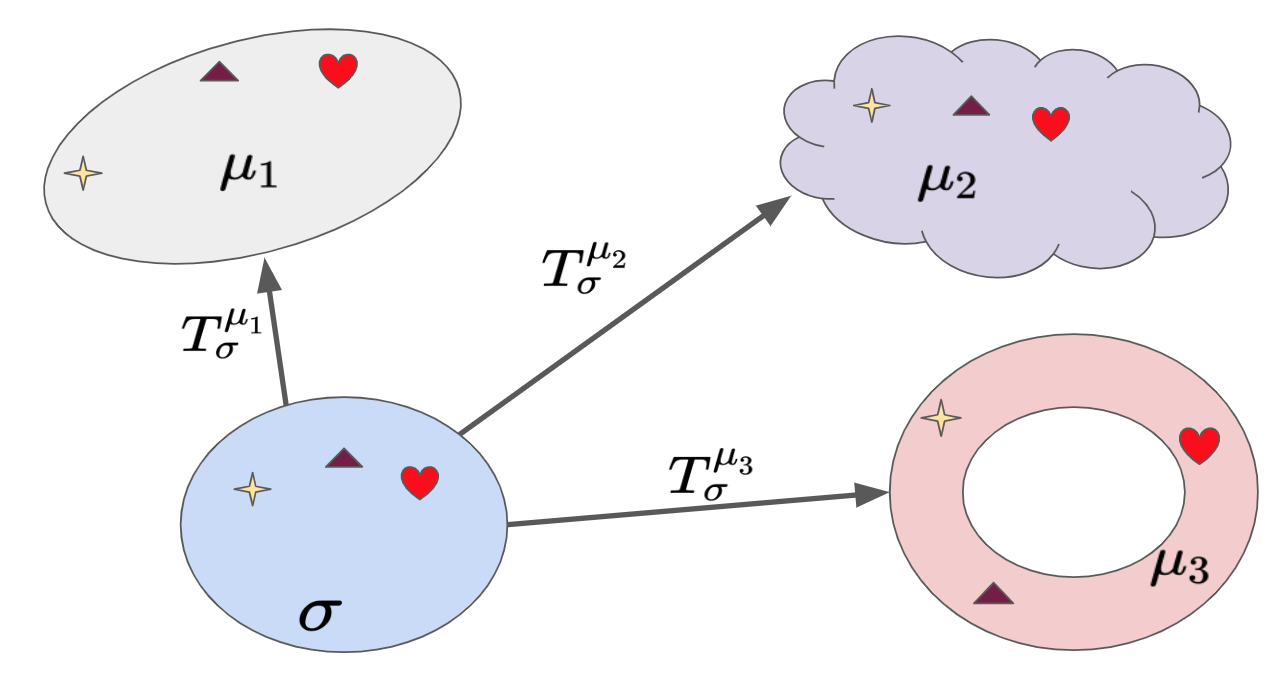}
        \caption{LOT corresponds to mix of registration of samples and interpolation of samples which registers datasets to fixed absolutely continuous distribution $\sigma$.}
        \label{fig:LOT_viz}
    \end{subfigure}
    \caption{LOT interpretations.}
    \label{fig:combined_lot}
\end{figure}

Specifically, let \(\{\mu_i\}_{i=1}^N\) be a collection of probability distributions on \(\mathbb{R}^d\), where each \(\mu_i\) represents an individual data point, and let \(\sigma\) be a fixed reference distribution on \(\mathbb{R}^d\). The key idea of LOT is to compute the optimal transport map \(T_\sigma^{\mu_i}\) from \(\sigma\) to \(\mu_i\), using the squared Euclidean distance as the cost function. The map \(T_\sigma^{\mu_i}\) belongs to the Hilbert space \(L^2(\sigma)\), which allows the embedding of the data point \(\mu_i\) into the space of square-integrable functions. A natural interpretation of this LOT embedding is applying the inverse-exponential map on the Wasserstein manifold at measure $\sigma$, sending each measure to the tangent space $L^2(\sigma)$. Mathematically, this embedding is expressed as
\begin{align}\label{intro:lot}
\mu_i \mapsto T_\sigma^{\mu_i}, & &\textnormal{ where } T_\sigma^{\mu_i}:= \arg\min_{T\in \Pi_\sigma^{\mu_i} } \int \|T(x)-x\|_2^2 d\sigma(x),
\end{align}
This mapping transforms each probability measure \(\mu_i\) into an element of \(L^2(\sigma)\), where we may use a wide range of classical machine learning algorithms, such as those for classification, dimension reduction, clustering, and data generation.  This interpretation of LOT is shown in \Cref{fig:lot_manifold}.

Another interpretation of LOT arises in the finite sample case.  Consider a collection of data sets $X_i\in\mathbb{R}^{n_i\times d}$ where the number of samples $n_i$ may vary.  A common problem is defining distances between these data sets that are independent of the order in which the samples are drawn.  This commonly arises when: the order of the samples collected is arbitrary, there's no a priori registration between nodes of two graphs, or the features of some data point are some unordered set.
Even when $n_i = n$ for all $i$, this requires learning a permutation invariant metric via the optimization scheme
\begin{align*}
    d(X_i, X_j) = \min_{P} \|X_i - P X_j\|.
\end{align*}
Obviously this metric does not induce an embedding space where one can perform downstream tasks on the data sets, and when the number of samples $n_i$ varies one must move to a more generalized notion of permutation matrices.

Using LOT immediately solves this problem.  When one draws $m$ points from $\sigma$ to push forward via $T_\sigma^{\mu_i}$ (or the finite sample point cloud $X_i$), the resulting LOT embedding is an $m\times d$ matrix $\widehat{T}_{X_i}$.  From here, one can define simple distances 
\begin{align*}
    d(X_i, X_j) = \|\widehat{T}_{X_i} - \widehat{T}_{X_j}\|,
\end{align*}
or even use the features $\widehat{T}_{X_i}$ directly for downstream tasks.  LOT performs a mix of registration of samples (when the number of samples are the same) and interpolation of samples (when the number of samples deviates) to register all data sets $X_i$ to a fixed absolutely continuous distribution $\sigma$.  While this registration may not be optimal, we argue that it is near optimal in a number of settings.  A visual for this interpretations is given in \Cref{fig:LOT_viz}.

Importantly, LOT proves especially valuable when the number of measure-valued data points is limited, as its nonparametric embedding can be readily applied without the need for extensive training. In contrast, parametric models often require large datasets to produce effective embeddings. Moreover, when the sample size of measure-valued data is sufficiently large to support parametric methods, LOT can serve as an excellent input that captures geometric information about each measure-valued data point.  Starting from robust LOT features allow parametric models to then further refine and optimize feature embeddings during training.

Given the importance of these tools for practical applications, we introduce \textbf{pyLOT}, a Python library designed to implement the general LOT framework. The pyLOT library provides users with a user-friendly platform for analyzing measure-valued data by offering the following key functionalities:

\begin{enumerate}
    \item \textbf{Embedding:} Measure-valued data can be embedded into an LOT space using either a pre-specified reference distribution $\sigma$ or a reference generated from the barycenter of the measure-valued data.  Here, we allow the ability to generate better reference distributions by iterating between approximating the barycenter of the measure-valued data and updating the reference to be the approximated barycenter.
    
    \item \textbf{Generation:} New measure-valued data can be generated by constructing a weighted barycenter in the LOT embedding space and pushing forward the reference measure \(\sigma\). This enables applications in generative modeling and simulation.
    
    \item \textbf{Dimension Reduction:} After embedding the measure-valued data into LOT space, dimension reduction techniques such as Linear Discriminant Analysis (LDA) or Principal Component Analysis (PCA) can be applied directly. This is particularly useful for visualizing high-dimensional measure data.
    
    \item \textbf{Classification:} Once LOT embeddings are obtained, standard classification algorithms can be applied for tasks such as K-Nearest Neighbors or support vector machines. The LOT space further ensures compatibility with more complicated downstream classifiers like random forests and neural networks.

\end{enumerate}

By embedding data into the Hilbert space \(L^2(\sigma)\), LOT opens up a wide range of possibilities for efficient analysis and processing of complex data. The pyLOT library makes these methods accessible to both researchers and practitioners, offering a versatile toolbox for modern machine learning tasks involving probability measures.

To demonstrate further efficacy of pyLOT to real-world datasets, we conduct a case study on a data set of high-resultion CT scans of primate molars, which are represented as point clouds. With this dataset, we showcase some of the functionality of pyLOT by performing dimensionality reduction, classification, and LOT barycenter generation after embedding the point clouds into an LOT space. This dataset serves as an ideal test case for evaluating the LOT framework due to its inherent complexity and small sample size.  In particular, we use the LOT embedding to perform a dimension reduction to visualize the clusters of teeth from different species, implement a few different supervised learning algorithms to classify new teeth, and use a random barycenter algorithm to generate new examples of teeth for each species.  We then compare the results of using the LOT embeddings with the results of using the pure Wasserstein approach, and we observe that the LOT approach has similar accuracy with a fraction of the computational cost, exactly as suggested by the theory.

\section{Background}

In this section, we will first discuss the details surrounding OT and LOT, and afterwards, we will discuss the details of the tooth dataset that we are using.

\subsection{Optimal Transport Preliminaries}

Consider the space of probability measures on $\R^n$, denoted as $\mathcal{P}(\R^n)$. A subset of this space, $\mathcal{P}_{\ac}(\R^n)$, consists of all probability measures that are absolutely continuous with respect to the Lebesgue measure. The Wasserstein space $W_2(\R^n)$ is composed of probability measures from $\mathcal{P}(\R^n)$ that have finite second moments, meaning that $\int_{\R^n} \|x\|^2 d\refe(x) < \infty$.  Optimal transport lends itself to two formulations: Kantorovich and Monge, where the Kantorovich is more general.  

In the Monge formulation, one is given two probability measures $\refe$ and $\temp$ and wants to find a map $S$ such that the pushforward of $\refe$ under $S$, denoted $S_{\sharp} \refe$, equals $\temp$. While many such maps might exist, optimal transport theory seeks the one that minimizes the cost of transporting $\refe$ to $\temp$. The cost function is typically taken to be the squared Euclidean distance, and the overall transport cost is given by
\begin{equation}\label{eq:cost}
    \int \| S(x) - x \|_2^2 d\refe(x).
\end{equation}
The solution to this minimization problem yields a unique, optimal transport map if it exists. In this case, the Wasserstein-2 distance between $\refe$ and $\temp$ can be expressed as
\begin{equation}\label{eq:Wasserstein_distance}
    W_2(\refe, \temp)^2 = \min_{S: S_\sharp \refe = \temp} \int \| S(x) - x \|_2^2 d\refe(x).
\end{equation}
This optimal transport distance defines a metric on $W_2(\R^n)$ and generalizes the notion of distance between probability measures. Additionally, optimal transport problems can be formulated for different norms beyond the Euclidean norm, such as $p$-norms, and can be extended to settings involving Riemannian manifolds \cite{brenier-1991, villani2008optimal, mccann-2001, ambrosio-2013}.

The Kantorovich formulation arises as a relaxation of the Monge problem because there may be cases when mass needs to split to actually solve the optimal transport problem.  In such cases, transport maps do not exist that solve the Monge problem.  If one allows mass to split, the optimal transport problem and Wasserstein metric can be stated as
\begin{equation}\label{eq:W2-metric}
    W_2(\refe,\temp) := \inf_{\pi \in \Gamma(\refe, \temp)} \left( \int_{\R^{2n}} \|x - y\|^2 d\pi(x, y) \right)^{\frac{1}{2}},
\end{equation}
where $\Gamma(\refe, \temp)$ refers to the set of all possible couplings between the measures $\refe$ and $\temp$. Formally, $\Gamma(\refe, \temp)$ is the collection of measures $\gamma \in \mathcal{P}(\R^{2n})$ such that $\gamma(A \times \R^n) = \refe(A)$ and $\gamma(\R^n \times A) = \temp(A)$ for every measurable set $A \subseteq \R^n$. These couplings capture joint distributions whose marginals correspond to $\refe$ and $\temp$.  When an optimal transport map exists, the optimal transport plan may be used to generate a plan and vice-versa.

Brenier's theorem \cite{brenier-1991} gives exact conditions of when a unique optimal transport map exists. We call this map ``the optimal transport from $\refe$ to $\temp$'' and denote it by $T_{\refe}^{\temp}$.  In particular, OT makes use of the following result:
\begin{theorem}[Brenier's theorem \cite{brenier-1991}]\label{Brenier} 
If $\refe \ll \lambda$ (the Lebesgue measure), the optimal transport map $T_{\refe}^{\temp}$ is uniquely defined as the gradient of a convex function $\varphi$, i.e.\ $T_{\refe}^{\temp}(x) = \nabla \varphi(x)$, where $\varphi$ is the unique convex function that satisfies $(\nabla \varphi)_\sharp \refe = \temp$. Uniqueness of $\varphi$ is up to an additive constant.
\end{theorem}

Importantly, since all real-world data is in the form of discrete measures, we must solve the Kantorovich problem with either a linear program or the Sinkhorn algorithm to produce an optimal transport plan.  The general procedure to generate transport plans is to construct the barycentric projection of the transport plans to convert them into maps.  In particular, if we have a discrete measures
\begin{align*}
    \refe = \sum_{i=1}^n \alpha_i \delta_{x_i}, \hspace{0.4cm} \temp = \sum_{j=1}^m \beta_j \delta_{y_j}
\end{align*}
supported at $\{x_i\}_{i=1}^n, \{y_j\}_{j=1}^m \subset \R^d$ respectively with $\alpha_i, \beta_j \geq 0$ such that $\sum_{j=1}^m \beta_j = \sum_{i=1}^n \alpha_i = 1$, then solving the discrete optimal transport problem will yield a coupling matrix $P_\refe^\temp = (p_{ij}^*) \in \R^{n \times m}$ that minimizes the square-Euclidean cost
\begin{align*}
    \sum_{i,j} p_{ij} \Vert x_i - y_j \Vert^2
\end{align*}
subject to the constraints
\[
    \sum_{i=1}^n p_{ij} = \beta_j \text{ for all } j\in[m], \qquad \sum_{j=1}^m p_{ij} = \alpha_)i \text{ for all } i\in[n].
\]
More concisely, if we define \(\mathbf 1_k\) to be the all-ones vector in \(\R^k\) and let $\alpha, \beta \in \R^k$ be the vectors corresponding to the weights $\alpha_i$ and $\beta_j$, then
\[
    P_\refe^\temp = \arg\min_{\substack{P\mathbf 1_m = \alpha \\ P^T\mathbf 1_n = \beta }} \sum_{i,j}p_{ij}\|x_i-y_j\|^2.
\]
The square root of the minimal value of this optimization is the 2-Wasserstein distance between \(\sigma\) and \(\mu\), and is denoted \(W_2(\sigma,\mu)\).  This optimization problem is a linear program and can be solved explicitly; however, the minimizer is not necessarily unique, in which case, it will suffice for our purposes to algorithmically choose a particular minimizer.  We can also solve the entropy-regularized optimization problem, where we do get a unique solution, but the solution may not be the optimal transport plan. After computing a minimal \(P\), we compute the optimal transport map \(T_\sigma^\mu\) by computing a Barycentric projection for each point \(x_i\) in the support of \(\refe\). Explicitly, we define \(\widetilde P\) to be the row-stochastic version of matrix \(P\).  Then we define \(T_\sigma^\mu\) at the \(n\) points in the support of \(\sigma\) by
\[
    T_\sigma^\mu (x_i) := \sum_{j=1}^m \widetilde P_{ij} y_j.
\]
Thus, \(T_\sigma^\mu(x_i)\) is precisely the average of the points in \(\mu\) weighted by the amount of mass sent to that point from \(x_i\) under the optimal transport plan \(P\). More detail can be found in, e.g., \cite{PC19}.  We point out one important fact, which is that in the case where \(m=n\) above, the optimal transport plan \(P\) will be exactly a permutation matrix.
More precisely, the optimal set will be the convex hull of permutation matrices (see, e.g., Proposition 2.1 in \cite{PC19}).

Most importantly, notice that when we work with point clouds, the same process for generating optimal transport maps works!

\subsection{Linearized Optimal Transport}

Following the work of Otto in \cite{otto2001geometry}, many connections have been drawn between optimal transport and differential geometry.
It is well known that the 2-Wasserstein distance \(W_2\) defines a metric on \(\mathcal P_2(\R^d)\), the space of probability measures with finite second moments.
Moreover, the metric space \(\mathbb W_2 = (\mathcal P_2(\R^d), W_2)\) can be interpreted as having a formal Riemannian structure, and so the language of differential geometry is often used to discuss properties of this ``Wasserstein manifold.''

For example, for a fixed reference \(\sigma\), one can interpret the embedding \(\mu\mapsto T_\sigma^\mu\) as a map from the Wasserstein manifold to the tangent space at \(\sigma\).
At a technical level, the logarithm map \(\log_\sigma^{\mathbb W_2}: \mathbb W_2 \to L^2(\sigma)\) is defined to be
\[
    \log_\sigma^{\mathbb W^2}(\mu) = T_\sigma^\mu - \id,
\]
but we omit the ``\(-\id\)'' in our calculations because the resulting LOT distance on the tangent space, defined as
\[
    W_{2,\sigma}^{LOT}(\mu_1,\mu_2) := \|T_\sigma^{\mu_1} - T_\sigma^{\mu_2}\|_{L^2(\sigma)} = \left(\int_{\R^d}\|T_\sigma^{\mu_1}(x) - T_\sigma^{\mu_2}(x)\|^2\,d\sigma(x)\right)^{1/2},
\]
is the same regardless of whether the \(\id\) component is included in the definition of the map to the space \(L^2(\sigma)\).
However, the map \(\mu\mapsto T_\sigma^\mu-\id\) gives a local diffeomorphism near \(\sigma\), which means the geometry of the Wasserstein manifold is locally preserved under this embedding. 
Subtracting the identity is a constant shift that does not affect the resulting geometry, and so we can say that the LOT embedding \(\mu\mapsto T_\sigma^\mu\) preserves the geometry of the Wasserstein manifold near \(\sigma\).
Using the language of differential geometry in this context is more for intuition than making precise statements, but it does help give a heuristic explanation for why the choice of \(\sigma\) can be important.
For example, if the point clouds \((\mu_i)_{i=1}^N\) are all close to each other, then choosing a nearby \(\sigma\) would likely yield better results in a classification algorithm because the point clouds' geometric relationship will be better preserved.

Theoretically, the reference measure is ideally chosen so that the resulting tangent space $L^2(\sigma)$ captures the local geometry of the measure-valued data well on the Wasserstein manifold.  When we assume structure of the underlying data-generating process, we can get more precise statements and results.  For example, \cite{khurana2022supervised} investigated conditions on the reference measure $\sigma$ and the underlying data-generating process $\{ h_\sharp \mu : h \in \mathcal{H} \}$ that guarantees that the LOT embedding is an isometry.  In particular, the isometry holds when the following compatibility condition 
\begin{align*}
    h \circ T_\sigma^\mu = T_{\sigma}^{h_\sharp \mu}
\end{align*}
holds; moreover, \cite{khurana2022supervised} showed that diffeomorphisms such as shears, shifts, and scaling and appropriate $\sigma$ ensure the compatibilty condition holds. Intuitively, the compatibility condition ensures the measure-valued data belongs to a flat portion of the Wasserstein manifold that is approximated by the tangent space at the reference measure $\sigma$, but this condition is sometimes too much regularity to ask for in practice and one can hope to get a better embedding by satisfying the $\epsilon$-compatible compatibility condition
\begin{align*}
    \Vert h \circ T_\sigma^\mu - T_{\sigma}^{h_\sharp \mu} \Vert_{L^2(\sigma)} < \epsilon.
\end{align*}
Since we usually do not have control of the data-generating process, we focus on choosing the best reference measure, and heuristically, the Wasserstein barycenter (or Fr{\'e}chet mean) of the measure-valued data seems to be a good candidate for capturing the local geometry of the data on the Wasserstein manifold.  The Wasserstein barycenter is computed using $\{\mu_i\}_{i=1}^n$ and a given weight vector $w$ (where $\sum_i w_i = 1$ and $w_i \geq 0$). Formally, the Wasserstein barycenter $\mu^*$ is defined as:
\begin{equation}
    \mu^* = \text{min} \sum_{i=1}^n w_i W_2^2(\mu, \mu_i)
\end{equation}
where $W_2$ is the 2-Wasserstein distance.  One algorithm to generate better embeddings is to update the reference as
\begin{align*}
    \sigma^{n+1} = \Big( \sum_{i=1}^n w_i T_{\sigma^n}^{\mu_i} \Big)_\sharp \sigma^{n}.
\end{align*}
It turns out \cite[Theorem 2.12]{kuang2019} that this algorithm will have $\sigma^{n+1}$ converge to the barycenter associated with $\{\mu_i\}_{i=1}^n$.  It seems, however, that no rates of convergence hold in general.  \cite{stromme2020barycenter} showed that if the data consists of Gaussian measures, the Polyak-\L{}ojasiewicz (PL) inequality holds and the first-order optimization scheme for barycenters as shown above converges linearly.  In general, \cite{stromme2020barycenter} showed that a integrated version of the PL-inequality holds but upgrading the inequality to a full PL-inequality requires a bound on the $L^\infty$-norm of the Radon-Nikodym derivative of all the iterates of the above scheme.

As LOT must be constructed from finite samples of each measure-valued data point $\mu$ in practice, concerns about statistical convergence must be addressed. Many results \cite{deb2021rates,pooladian2021} give concentration inequalities with nonasymptotic rates for when the barycentric projection of the optimal transport plan either computed from solving a linear program or by using the Sinkhorn algorithm.  The general form of the convergence rates are of the form $\approx k^{-1/d} + m^{-1/2}$, where $k$ is the sample size associated with the measure-valued data point $\mu$, $m$ is the sample size associated with the reference measure $\sigma$, and $d$ corresponds to the dimension.  When applying LOT, preserving the geometry in LOT space when using finite-sample approximations is of upmost importance and easily seen as a consequence of the convergence rates.  \cite{cloninger2023lotwassmap} showed that the finite-sampled approximated LOT distance concentrates around the true Wasserstein distance of the meausre-valued data provided that the data and reference measures satisfy $\epsilon$-compatibility.  These statistical rates show that LOT still works with varying amounts of samples from each individual measure-valued data point albeit at the cost of slower rates of convergence.  In practice, LOT acts as a framework for different sized point clouds and allows comparison between those different point clouds in a theoretically-supported manner.

In this paper, we will focus more on the experimental results of performing ML algorithms after using the LOT embedding. However, it is important to note that there are many regularity results that guarantee nice properties for this embedding. For example, \cite{khurana2022supervised} showed that if the data-generating process and reference measure satisfies the compatibilty condition and the data is ``separable'' in Wasserstein space, then the data is linearly separable in the LOT embedding space.  For unsupervised learning methods, note that \cite{cloninger2023lotwassmap} showed that if the measure-valued data is almost-compatible and the number of samples from the each measure-valued data point as well as the reference measure is large, then applying multi-dimensional scaling (MDS) to the measure-valued data will recover low dimensional structures inherent to the data with high probability.

\subsection{Related Work}

\subsubsection{Point-cloud data analysis methods}

The LOT embedding method is not the first technique to analyze point cloud-valued data. However, there have been significant advances in the development of methods that address these challenges, allowing machine learning to operate on point-cloud-valued data. Graph neural networks \cite{wang2019} treat point-cloud data as graphs and propagate information through this graph to learn features useful for classification or segmentation. 
DeepSets \cite{zaheer2017deepsets} provides another neural network architecture that is permutation invariant to the inputs which allow it to work for point-cloud data quite easily.  PointNet \cite{qi2017} and PointNet++ \cite{qi2017pointnetplus} introduce a deep learning architecture that consumes raw point-clouds and outputs classification and segmentation results. One can also create an encoder that learns a compact latent representation of the point-cloud, and a decoder that attempts to reconstruct the original point-cloud from this latent space \cite{achlioptas2018learning}. To extract topological features from point-clouds, one uses topological data analysis (TDA) methods \cite{cao2022} like persistent homology. To learn more geometric features of the point-cloud data, one can generate a graph from the point-cloud and consider spectral properties of the graph. In particular, \cite{robertson2024} generalizes Wasserstein distance to connection graphs and explores its use in supervised and unsupervised learning on point-cloud-valued data. As permutation invariance is needed for point-cloud data analysis, LOT indeed exhibits this permutation invariance since it uses optimal transport to generate the embeddings.

Recently, \cite{werenski2024} explored the \textit{synthesis} and \textit{analysis} problem (given by barycentric coding models) associated with the Wasserstein barycenter problem and related it to the linearized Wasserstein barycentric coding model. They showed that if compatibility holds for all the probability measures involved, then solving the linearized barycenter synthesis problem will also solve the regular Wasserstein barycenter synthesis problem.  We implement the method that solves linearized Wasserstein barycenter synthesis problem, but we call it LOT barycenters.  We further implement iterative embeddings, where we iterate between generating an LOT barycenter and updating the reference measure for LOT embeddings to be the LOT barycenter, and see that the LOT barycenter after a few iterations becomes much closer to the true barycenter.

\subsubsection{Tooth dataset}

The tooth dataset that we use is a collection of 3D scans of teeth, originally utilized for morphological and geometric analysis of biological structures. This dataset primarily contains detailed point-cloud representations of teeth, which capture both the shape and size of each tooth in high resolution. It's main use in research studies has been to explore anatomical variations, evolutionary relationships, and geometric shape analysis across different species. A notable characteristic of the dataset is its application to comparative biology, where it has enabled the study of dental morphology and its implications for taxonomic classification, functional morphology, and evolutionary adaptation. Each tooth in the dataset is represented as a point-cloud or mesh, allowing for advanced computational techniques to quantify geometric similarity, perform shape analysis, and classify dental structures.

The dataset was first studied by \cite{Boyer2011} to automatically quantify geometric similarity in anatomical structures, particularly focusing on the analysis of 3D tooth shapes. Subsequently, \cite{Gao2021} introduced a new method of analyzing datasets represented as fiber bundles, applying diffusion maps to capture the geometric structure of the data. \cite{StClair2016} studied molar shape and size variation across different primate species, using the Tooth Dataset for its detailed 3D reconstructions of dental structures. The work contributed to understanding evolutionary changes in dental morphology.

Our case study is conducted on this curated dataset comprised of high-resolution CT scans of $N = 58$ mandibular molars from two primate suborders. The genera analyzed include Microcebus ($N = 11$), Mirza ($N = 5$), Saimiri ($N = 9$), and Tarsius ($N = 33$). Each tooth is represented as a pair $(\mu_i, y_i)$, where $\mu_i \subset \mathbb{R}^3$ is a point cloud consisting of over 5000 points in three-dimensional space, and $y_i$ denotes the corresponding genus label.

\section{pyLOT Functionality with Case Studies}

Our repository pyLOT acts as a lightweight implementation of the LOT framework and applications of the LOT embedding, thereof.  The fundamental module of pyLOT is the embedding module, which can embed a list of discrete measures into an LOT space with respect to a discrete reference measure.  Once all these embeddings are calculated, one can apply dimensionality reduction, classify data, and generate new measure-valued data.  We will explain the detailed functionality of each submodule (embedding, dimension reduction, classification, and LOT barycenter generation) and show how the module is applied to the tooth dataset described above.

\subsection{Preprocessing via LOT Embedding}

The \texttt{LOTEmbedding} submodule implements the Linearized Optimal Transport (LOT) embedding for point cloud data. It provides tools for embedding complex geometric structures, such as 3D point clouds of primate teeth, into a Hilbert space where linear methods can be applied. A key feature of this submodule is the iterative computation of barycenters to refine the reference measure, enhancing the representation of the data.

Given a reference point cloud $\mathbf{x}_r \in \mathbb{R}^{m \times d}$ and a target point cloud $\mathbf{x}_t \in \mathbb{R}^{n \times d}$, the LOT embedding maps $\mathbf{x}_t$ into the space of $\mathbf{x}_r$ by computing the optimal transport (OT) map between them.  The OT map is computed by solving the following optimization problem:
\begin{align} 
\min_{\mathbf{G} \in \mathbb{R}^{m \times n}} \quad & \sum_{i=1}^{m} \sum_{j=1}^{n} G_{ij} M_{ij} \ \text{subject to} \quad & \mathbf{G} \mathbf{1}_n = \mathbf{a}, \ & \mathbf{G}^\top \mathbf{1}m = \mathbf{b}, \ & G{ij} \geq 0, 
\end{align}
where $\mathbf{G}$ is the transport plan, $M_{ij} = \Vert (\mathbf{x}_{r})_i - (\mathbf{x}_{t})_{j} \Vert_2$ is the cost matrix based on Euclidean distances (or user-defined cost matrix), $\mathbf{a} \in \Delta^m$ and $\mathbf{b} \in \Delta^n$ are the mass distributions (defaulting to uniform distributions if not provided), $\mathbf{1}_n$ and $\mathbf{1}_m$ are vectors of ones of appropriate dimensions, and $\Delta^k$ denotes the $k$-dimensional probability simplex.

As this problem has a linear cost and constraints, we can view this problem as a linear program and can solve it as such to get an optimal transport plan $\mathbf{G}$.  As the sample sizes $m$ and $n$ get large, however, solving the linear program becomes computationally prohibitive.  For computational efficiency, the Sinkhorn algorithm can be used to solve the entropically-regularized discrete optimal transport problem, which yields an approximation to the OT map:
\begin{equation} \mathbf{G}^\lambda = \operatorname{Sinkhorn}(\mathbf{a}, \mathbf{b}, M, \lambda), \end{equation}
where $\lambda > 0$ is the regularization parameter.

Regardless of whether the problem is solved by linear programming or the Sinkhorn algorithm, the optimal transport plan needs to be transformed back into a map.  Barycentric projection $\mathbf{T} \in \mathbb{R}^{m \times d}$ maps the target point cloud into the reference space
\begin{equation} \mathbf{T} = \mathbf{G}_{\text{stochastic}} \mathbf{x}_t, \end{equation}
where $\mathbf{G}_{\text{stochastic}}$ is the row-normalized transport plan
\begin{equation} \mathbf{G}_{\text{stochastic}} = \operatorname{diag}\left( \mathbf{G} \mathbf{1}_n + \epsilon \right)^{-1} \mathbf{G}.
\end{equation} 
Here, $\epsilon$ is a small constant to ensure numerical stability.

To embed a collection of target point clouds $\{ \mathbf{x}_{t}^{(k)} \}_{k=1}^K$, the submodule iteratively computes the barycentric projections for each transport plan
\begin{equation} \mathbf{T}^{(k)} = \mathbf{G}_{\text{stochastic}}^{(k)} \mathbf{x}_{t}^{(k)}, \quad \text{for } k = 1, \dots, K, \end{equation}
where $\mathbf{G}_{\text{stochastic}}^{(k)}$ corresponds to the optimal transport plan from solving the problem between the reference point cloud and the target point cloud $\mathbf{x}_t^{(k)}$.  These embeddings $\mathbf{T}^{(k)}$ are then flattened and stacked to form the dataset for further analysis.  The function used to generate embeddings has function signature \texttt{LOTEmbedding.embed\_point\_clouds(xr, xt\_lst, r\_mass=None, xt\_masses=None, sinkhorn=False, lambd=1, normalize\_T=False)}, where the function has default values as listed.  The inputs and outputs of \texttt{LOTEmbedding.embed\_point\_clouds} are shown as
\begin{itemize}
    \item \texttt{xr (ndarray)}: The reference point cloud to which other point clouds will be compared. Shape: $(m, dr)$.
    \item \texttt{xt\_lst (ndarray)}: A list or array of target point clouds to embed. Shape: (\texttt{num\_point\_clouds}, $n$, $dt$).
    \item \texttt{r\_mass (ndarray, optional)}: An array assigning mass to each point in the reference point cloud \texttt{xr}. Default is \texttt{None} (uniform mass).
    \item \texttt{xt\_masses (ndarray, optional)}: A list of arrays assigning mass to each point in the corresponding target point cloud in \texttt{xt\_lst}. Default is \texttt{None} (uniform mass).
    \item \texttt{sinkhorn (bool, optional)}: Indicates whether to use the Sinkhorn algorithm for optimal transport (OT) calculation. Default is \texttt{False} (use exact OT).
    \item \texttt{lambd (float, optional)}: Regularization parameter for the Sinkhorn algorithm. Default is 1.
    \item \texttt{normalize\_T (bool, optional)}: Determines whether to normalize the transport maps by the square root of the number of reference points. Default is \texttt{False}.
\end{itemize}
The output is
\begin{itemize}
    \item \texttt{pclouds (ndarray)}: A stacked array of flattened LOT embeddings of the target point clouds. Shape: (\texttt{num\_point\_clouds}, $m \cdot dt$).
\end{itemize}

\subsubsection{LOT embedding of Tooth Dataset}

\begin{figure}[h!]
    \centering
        \begin{tabular}{cc}
        \includegraphics[scale = 0.4]{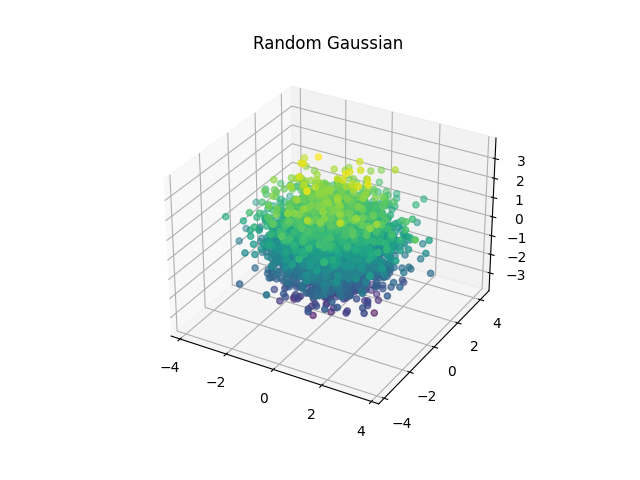} &
        \includegraphics[scale = 0.4]{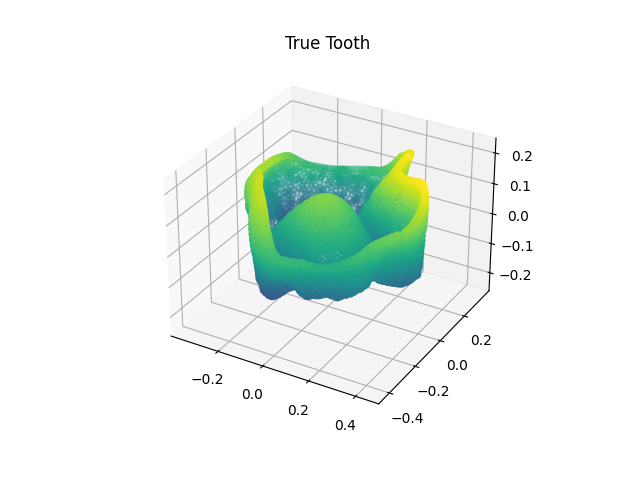} \\
        \includegraphics[scale = 0.4]{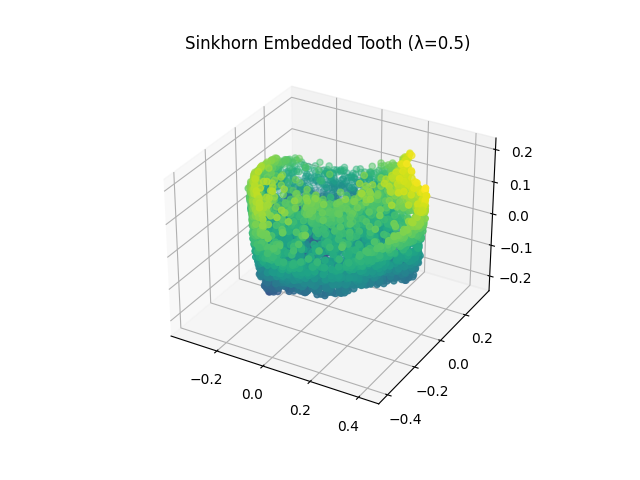} &
        \includegraphics[scale = 0.4]{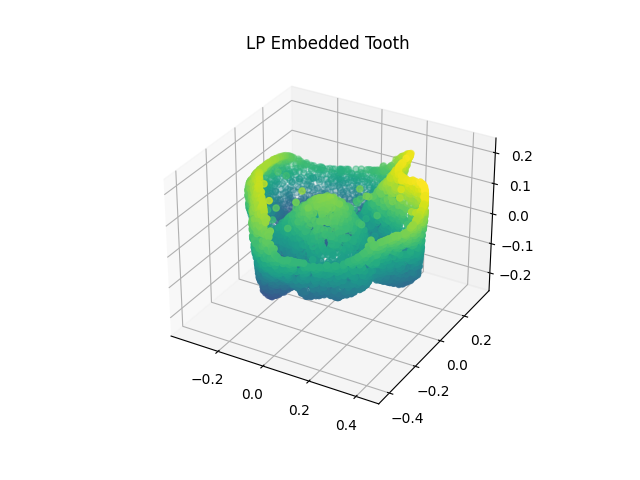}
        \end{tabular}
    \caption{Tooth Comparison}
    \label{fig:tooth_comparison}
\end{figure}

By applying the \texttt{LOTEmbedding} submodule to 3D point clouds of primate teeth, we effectively capture morphological variations in the embedding, which enables analyses such as classification or clustering in the embedded space.  To see this module in practice, we embedded a tooth point-cloud into the LOT embedding space in \Cref{fig:tooth_comparison}, where we use a sample from a random Gaussian as the reference measure.  In particular, we applied LOT to embed each tooth point cloud into a $5000 \times 3$ matrix, constructing the dataset ${\mathbf{T}^{(k)}}_{j=1}^N$ used in subsequent analyses. We generate a reference point cloud $\mathbf{x}_r \in \mathbb{R}^{5000 \times 3}$ using a Gaussian distribution. As the embedding is a stacked approximation map from the samples of the Gaussian to the point cloud, we reshape the embedding to get the point clouds seen in \Cref{fig:tooth_comparison}.  For each tooth point cloud, we compute the embedding using two methods: Linear Programming (LP) solution and Sinkhorn algorithm. This associated code is in Listing \ref{code:lotEmbed}.

\begin{lstlisting}[language=Python, caption=Code to embed tooth samples., label={code:lotEmbed}]
from pyLOT.embed import LOTEmbedding
xt_list, T_labels = load_fps() # load tooth dataset with appropriate labels
xr = np.random.normal(0, 1, size=(5000, 3)) # set reference measure with 5000 points

# Compute LOT Sinkhorn embeddings with small lambda
T_embeddings_sh = LOTEmbedding.embed_point_clouds(xr, 
                                            xt_list,
                                            xt_masses=None, # allows masses if needed.  Defaults to None (i.e. assume uniform mass)
                                            sinkhorn=True, 
                                            lambd=0.05)

# Compute LOT embeddings
T_embeddings = LOTEmbedding.embed_point_clouds(xr, 
                                            xt_list,
                                            sinkhorn=False)
\end{lstlisting}

Note that the embedding indeed captures valuable information about the shape of the tooth.  Moreover, as the Sinkhorn algorithm slightly blurs the target measure, we see a fuzzier approximation of the true tooth.  In contrast, the embedding generated from solving the linear program preserves the fine-scale morphological details of the teeth such as the edges, which are more conducive to subsequent analytical tasks across primate genera.  This initial inspection suggests that the linear program solution to the optimal transport problem may be better suited to data with point-clouds that have sharp differentiating features.  On the other hand, if the point-clouds do not need sharp features to be differentiated, the Sinkhorn solution can suffice.  For the remainder of the paper, the embeddings used for the subsequent sections are generated with linear programming.

\subsection{Dimension Reduction}

The \texttt{LOTDimensionalityReduction} submodule provides tools for reducing the dimensionality of Linearized Optimal Transport (LOT) embeddings derived from discrete measure-valued (or point-cloud-valued) data. It includes methods for balancing imbalanced datasets and applying dimensionality reduction techniques such as Linear Discriminant Analysis (LDA) and Principal Component Analysis (PCA). These methods facilitate the analysis and visualization of high-dimensional LOT embeddings.

As imbalanced datasets can bias the results of dimensionality reduction and classification algorithms, the submodule includes a data balancing method using random oversampling of minority classes.

Given a dataset of LOT embeddings $\mathbf{T} \in \mathbb{R}^{N \times D}$ and corresponding labels $\mathbf{y} \in {1, \dots, C}^N$, the data balancing process aims to equalize the number of samples in each class. In particular, we first determine the number of samples $N_c$ in each class $c \in {1, \dots, C}$. Now for each class, we compute the required number of samples to match the majority class. Finally, to balance the dataset, we randomly duplicate samples from minority classes until all classes have $N_{\text{max}}$ samples, where $N_{\text{max}} = \max{N_1, \dots, N_C}$. This results in a balanced dataset $(\mathbf{T}_{\text{bal}}, \mathbf{y}_{\text{bal}})$ with equal representation from all classes.

\subsubsection{Linear Discriminant Analysis (LDA) Deduction}
LDA is a supervised dimensionality reduction technique that projects data onto a lower-dimensional space while maximizing class separability. Using the balanced dataset $(\mathbf{T}_{\text{bal}}, \mathbf{y}_{\text{bal}})$, LDA computes a transformation matrix $\mathbf{W}_{\text{LDA}} \in \mathbb{R}^{D \times (C - 1)}$ by solving
\begin{equation}
    \mathbf{W}_{\text{LDA}} = \arg\max{\mathbf{W}} \frac{\left| \mathbf{W}^\top \mathbf{S}_B \mathbf{W} \right|}{\left| \mathbf{W}^\top \mathbf{S}_W \mathbf{W} \right|},
\end{equation}
where $\mathbf{S}_B$ is the between-class scatter matrix and $\mathbf{S}_W$ is the within-class scatter matrix. The transformed data is then given by
\begin{equation}
    \mathbf{Z}_{\text{LDA}} = \mathbf{T}_{\text{bal}} \mathbf{W}_{\text{LDA}} \in \mathbb{R}^{N_{\text{bal}} \times (C - 1)}.
\end{equation}
The LDA-transformed embeddings $\mathbf{Z}_{\text{LDA}}$ can be used for classification or visualization, providing a reduced representation that emphasizes class distinctions.  The code associated to computing the LDA reduction has signature \texttt{LOTDimensionalityReduction.lda\_reduction(pclouds, labels, n\_components=3)} with inputs given by
\begin{itemize}
    \item \texttt{pclouds (np.array)}: A 2D array where each row represents a point cloud (data point). Shape: (\texttt{number\_of\_point\_clouds}, \texttt{number\_of\_features}).
    \item \texttt{labels (np.array)}: A 1D array of labels corresponding to the point clouds in \texttt{pclouds}. Shape: (\texttt{number\_of\_point\_clouds},).
    \item \texttt{n\_components (int, optional)}: The number of components to retain after performing LDA. Default is 3.
\end{itemize}
The outputs are
\begin{itemize}
    \item \texttt{T\_lda (np.array)}: A 2D array representing the point clouds transformed into the space defined by the LDA components. Shape: (\texttt{number\_of\_point\_clouds\_balanced}, \texttt{n\_components}).
    \item \texttt{labels\_balanced (np.array)}: A 1D array of labels corresponding to the balanced point clouds. Shape: (\texttt{number\_of\_point\_clouds\_balanced},).
\end{itemize}

\subsubsection{Principal Component Analysis (PCA) Reduction}

PCA is an unsupervised dimensionality reduction technique that identifies the directions (principal components) capturing the most variance in the data. Computing PCA involves first centering the data by
\begin{equation} 
    \mathbf{T}_{\text{centered}} = \mathbf{T}_{\text{bal}} - \mathbf{\mu}, \quad \text{where} \quad \mathbf{\mu} = \frac{1}{N_{\text{bal}}} \sum_{i=1}^{N_{\text{bal}}} (\mathbf{T}_{\text{bal}})_{i}.
\end{equation} 
Next, we compute the covariance matrix
\begin{equation} 
    \mathbf{C} = \frac{1}{N_{\text{bal}} - 1} \mathbf{T}_{\text{centered}}^\top \mathbf{T}_{\text{centered}} \in \R^{D \times D}.
\end{equation} 
and its eigenvalue decomposition \begin{equation} 
    \mathbf{C} = \mathbf{V} \mathbf{\Lambda} \mathbf{V}^\top,
\end{equation} 
where $\mathbf{V}$ contains the eigenvectors (principal components), and $\mathbf{\Lambda}$ is a diagonal matrix of eigenvalues. Finally, we project onto principal components
\begin{equation} 
    \mathbf{Z}_{\text{PCA}} = \mathbf{T}_{\text{centered}} \mathbf{V}_k \in \R^{N_{\text{bal}} \times k}, 
\end{equation} 
where $\mathbf{V}_k$ contains the top $k$ principal components corresponding to the largest eigenvalues.

The PCA-transformed embeddings $\mathbf{Z}_{\text{PCA}}$ provide a reduced-dimensional representation capturing the most significant variance in the data, useful for visualization and exploratory data analysis.  Moreover, one can perform calculations, such as the LOT barycenter, in the reduced space and apply the pseudoinverse of the projection $V_k$ to get a de-noised version of the LOT barycenter, which could be more stable compared to the regular LOT barycenter.  The code associated to computing the PCA reduction has signature \texttt{LOTDimensionalityReduction.pca\_reduction(pclouds, labels)} with inputs given by
\begin{itemize}
    \item \texttt{pclouds (np.array)}: A 2D array where each row represents a point cloud (data point). Shape: (\texttt{number\_of\_point\_clouds}, \texttt{number\_of\_features}).
    \item \texttt{labels (np.array)}: A 1D array of labels corresponding to the point clouds in \texttt{pclouds}. Shape: (\texttt{number\_of\_point\_clouds},).
\end{itemize}
The outputs are
\begin{itemize}
    \item \texttt{U (np.array)}: The left singular vectors (principal components) of the centered data. Shape: (\texttt{number\_of\_point\_clouds\_balanced}, \texttt{min(number\_of\_point\_clouds\_balanced, number\_of\_features)}).
    \item \texttt{S (np.array)}: The singular values corresponding to the principal components. Shape: (\texttt{min(number\_of\_point\_clouds\_balanced, number\_of\_features)},).
    \item \texttt{Vh (np.array)}: The right singular vectors, representing the principal axes in the feature space. Shape: (\texttt{min(number\_of\_point\_clouds\_balanced, number\_of\_features)}, \texttt{number\_of\_features}).
    \item \texttt{labels\_balanced (np.array)}: A 1D array of labels corresponding to the balanced point clouds. Shape: (\texttt{number\_of\_point\_clouds\_balanced},).
\end{itemize}

\subsubsection{Dimensionality reduction applied to Tooth Dataset}

\begin{figure}[h!]
    \centering
        \begin{tabular}{cc}
        \includegraphics[scale = 0.4]{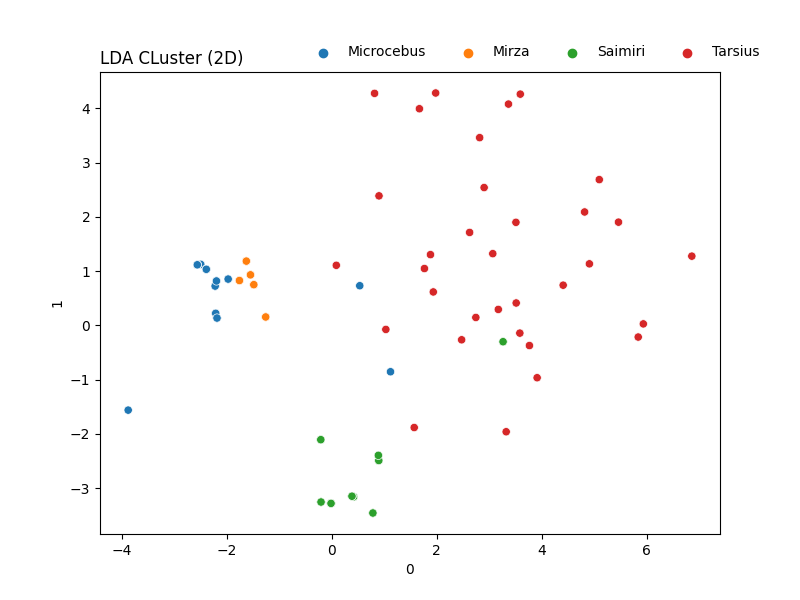} &
        \includegraphics[scale = 0.4]{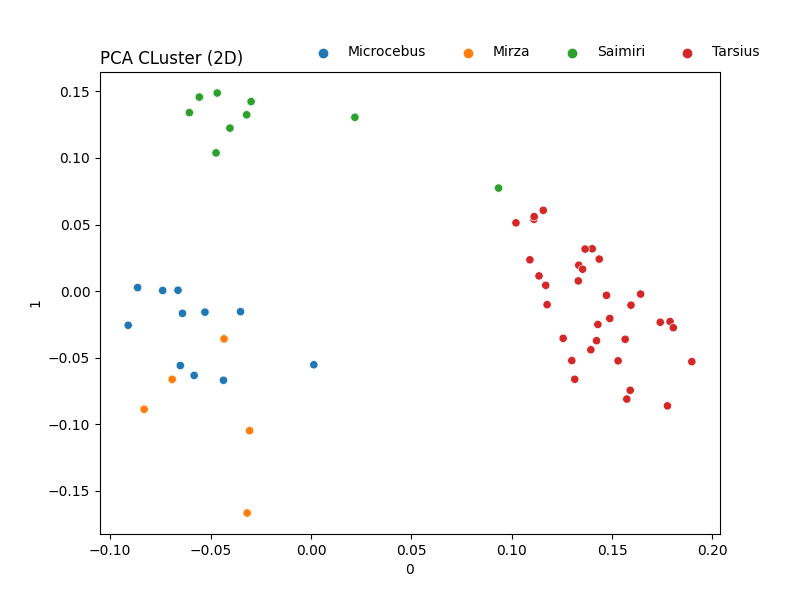}\\
        \includegraphics[scale = 0.4]{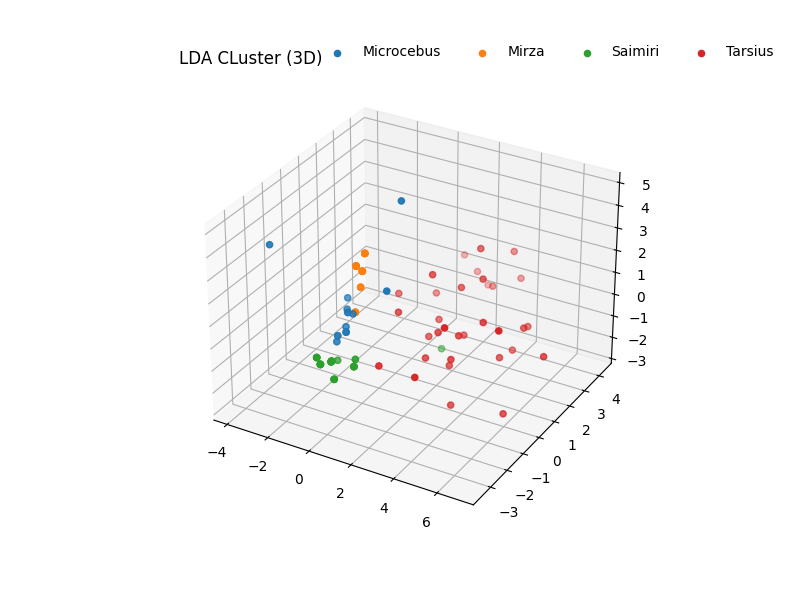} &
        \includegraphics[scale = 0.4]{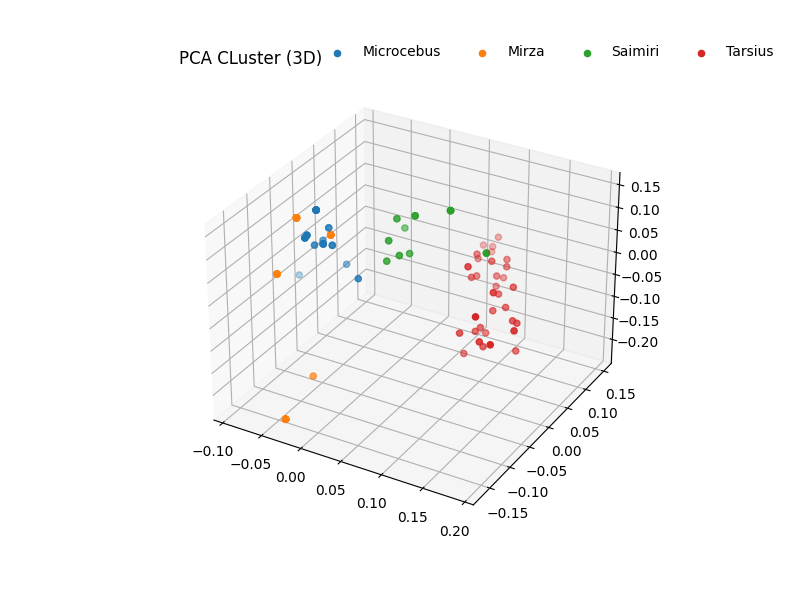}
        \end{tabular}
    \caption{PCA VS. LDA Cluster}
    \label{fig:cluster}
\end{figure}

In our experiment, we first addressed class imbalance through oversampling to ensure fair representation across all genera. We then applied both PCA and LDA to visualize the expected clusters as shown in \Cref{fig:cluster}. Theoretically, LDA should provide better class separation for classification tasks, given its supervised nature and focus on maximizing between-class variance while minimizing within-class variance.  The code to reproduce the tests here are given in Listing \ref{code:reduction}

\begin{lstlisting}[language=Python, caption=Code for PCA and LDA reduction., label={code:reduction}]
# run LDA for T_embeddings
T_lda, labels_balanced = LOTDimensionalityReduction.lda_reduction(T_embeddings, T_labels, n_components=3)

# run PCA for T_embeddings
U, S, Vh, labels_balanced = LOTDimensionalityReduction.pca_reduction(T_embeddings, T_labels)
\end{lstlisting}

Contrary to our initial expectations, however, the PCA plots exhibited more distinguishable boundaries between classes compared to the LDA plots. Both plots revealed intersections between Microcebus (Blue) and Mirza (Yellow) data points, indicating a challenge in differentiating these genera. However, the LDA plot presented an additional complication: a point of intersection between Saimiri (Green) and Tarsius (Red) data, which was not observed in the PCA plot. 

\subsection{Classification}

The \texttt{LOTClassifier} submodule provides tools for selecting and evaluating classifiers on the embeddings obtained from Linearized Optimal Transport (LOT) of point cloud data. It automates the process of choosing the best-performing classifier based on cross-validation accuracy and offers detailed performance metrics, facilitating effective classification tasks on high-dimensional data.

The primary function of this submodule is to determine the best classifier from a predefined set by evaluating their performance on the training data using cross-validation and then testing them on unseen data.  Given LOT embedded training data $(T_{\text{train}}, y_{\text{train}})$ and test data $(T_{\text{test}}, y_{\text{test}})$, the submodule first initializes classifiers ranging from $K$-nearest neighbors (KNN) with $k = 3$ neighbors, support vector machines (SVM) with linear kernel, and SVMs with a radial basis function (RBF) kernel.

For each classifier, 5-fold cross-validation is performed on the training set. The mean cross-validated accuracy $\bar{A}_{\text{cv}}$ is computed by
\begin{equation} 
    \bar{A}_{\text{cv}} = \frac{1}{5} \sum_{i=1}^{5} (A_{\text{cv}})_{i}, 
\end{equation} 
where $(A_{\text{cv}})_{i}$ is the accuracy in the $i$-th fold. Each classifier is then trained on the entire training set, and the trained model is evaluated on the test set to compute the test accuracy $A_{\text{test}}$.  Validation and test accuracies are reported for each classifier and a detailed classification report is generated, including metrics such as precision, recall, and F1-score for each class. The classifier with the highest mean cross-validated accuracy $\bar{A}_{\text{cv}}$ is selected as the best classifier.  The code associated to computing the best classifier has signature \texttt{LOTClassifier.get\_best\_classifier(X\_train, y\_train, X\_test, y\_test)} with inputs given by
\begin{itemize}
    \item \texttt{X\_train (np.array)}: A 2D array representing the training data features. Shape: (\texttt{number\_of\_training\_samples}, \texttt{number\_of\_features}).
    \item \texttt{y\_train (np.array)}: A 1D array of labels corresponding to the training data features. Shape: (\texttt{number\_of\_training\_samples},).
    \item \texttt{X\_test (np.array)}: A 2D array representing the test data features. Shape: (\texttt{number\_of\_test\_samples}, \texttt{number\_of\_features}).
    \item \texttt{y\_test (np.array)}: A 1D array of labels corresponding to the test data features. Shape: (\texttt{number\_of\_test\_samples},).
\end{itemize}
The outputs are
\begin{itemize}
    \item \texttt{best\_classifier (sklearn classifier)}: The classifier with the highest cross-validated accuracy on the training set. This could be one of the following:
    \begin{itemize}
        \item \texttt{KNeighborsClassifier}: A K-Nearest Neighbors classifier with 3 neighbors.
        \item \texttt{SVC (Linear)}: A Support Vector Classifier with a linear kernel.
        \item \texttt{SVC (RBF)}: A Support Vector Classifier with an RBF (Gaussian) kernel.
    \end{itemize}
\end{itemize}

\subsubsection{Classification of Tooth Dataset}

Using the embedded dataset ${T_\sigma^{\mu_i}}_{i=1}^N$ obtained through LOT, we applied traditional machine learning algorithms for classification. Specifically, we applied k-Nearest Neighbors (KNN), Linear Support Vector Machine (Linear SVM), and Radial Basis Function Support Vector Machine (RBF SVM).

To address class imbalance in our dataset, we first performed a train-test split on the original dataset. Then we applied random oversampling with replacement separately within both the training and testing sets. This process involved randomly duplicating existing samples from minority classes until all classes had an equal number of instances within each set. By applying oversampling independently to both training and testing sets after the initial split, we ensured that each class is equally represented during both the learning process and evaluation, while avoiding potential data leakage between the training and testing sets.  The code associated with the classification is given in Listing \ref{code:classify}.
\begin{lstlisting}[language=Python, caption=Code for classification of teeth., label={code:classify}]
# assume data has been split and oversampled to create X_train, y_train, X_test, y_test

# Train the classifier on the data
best_clf = LOTClassifier.get_best_classifier(X_train, y_train, X_test, y_test)
\end{lstlisting}

After oversampling, our final dataset consisted of 108 samples in the training set and 24 samples in the testing set.  The cross-validation performance of these models is summarized in \Cref{tab:val_performance} whilst the test performance of the models is summarized in \Cref{tab:test_performance}.

\begin{table}[H]
\centering
\begin{tabular}{|l|c|}
\hline
\textbf{Model} & \textbf{Val. Accuracy} \\
\hline
KNN & 0.990 \\
\textbf{Linear SVM} & \textbf{1.00} \\
RBF SVM & 0.897 \\
\hline
\end{tabular}
\vspace{0.3cm}
\caption{Validation Performance}
\label{tab:val_performance}
\end{table}

All models demonstrate exceptionally high performance, suggesting that the LOT embeddings provide highly discriminative features for classification. The perfect scores for Linear SVM on the test set may be attributed to the very small dataset used in this example. Further validation on larger, more diverse datasets would be beneficial to confirm the generalizability of these models and the effectiveness of the LOT embedding approach.

\begin{table}[H]
\centering
\begin{tabular}{|l|c|c|c|c|}
\hline
\textbf{Model} & \textbf{Test Accuracy} & \textbf{Avg. Precision} & \textbf{Avg. Recall} & \textbf{Avg. F1-score} \\
\hline
KNN & 0.875 & 0.92 & 0.88 & 0.87 \\
\textbf{Linear SVM} & \textbf{1.00} & \textbf{1.00} & \textbf{1.00} & \textbf{1.00} \\
RBF SVM & 0.875 & 0.92 & 0.88 & 0.87 \\
\hline
\end{tabular}
\vspace{0.3cm}
\caption{Test Performance}
\label{tab:test_performance}
\end{table}

\subsection{LOT Barycenter Generation}

Recall the Wasserstein barycenter as solving
\begin{equation}
    \mu^* = \text{argmin} \sum_{i=1}^n w_i W_2^2(\mu, \mu_i)
\end{equation}
where $\{\mu_i\}_{i=1}^n$ are the measures we are trying to find the barycenter with respect to and $w$ is a weight vector $w$ (where $\sum_i w_i = 1$ and $w_i \geq 0$).  The \texttt{LOTBarycenter} submodule provides methods for generating barycenters (weighted averages) of point clouds and general measure-valued data.  The main idea of generating barycenters with LOT embeddings is to first generate the LOT embeddings $\{\mathbf{T}^{(k)}\}_{k=1}^N$ of your data $\{\mathbf{x}_t^{(k)}\}_{k=1}^N$ with respect to a reference $\mathbf{x}_r$, then you generate a Wasserstein barycenter approximation by performing the simple $L^2(\mathbf{x}_r)$ barycenter/average
\begin{align*}
    \mathbf{x}^* = \Big( \sum_{k=1}^N w_k \mathbf{T}^{(k)} \Big)_{\sharp} \mathbf{x}_r,
\end{align*}
where $w$ is a weight vector with $\sum_{k=1}^N w_k = 1$ and $w_k \geq 0$.  We call $\mathbf{x}^*$ the LOT barycenter.

These LOT barycenters can be computed within the same class, between different classes, or from arbitrary sets of point-clouds/measures using specified weights. Once the LOT barycenters are generated, one needs to simply reshape the data to return to the space of discrete measures. Moreover, if one wishes to have a de-noised version of the LOT barycenter, they can apply dimensionality reduction such as PCA, compute the barycenter in the reduced space, and apply the pseudoinverse of the dimensionality reduction projection matrix to get a de-noised version of the LOT barycenter. Many applications require a Wasserstein barycenter for representing average shapes or generating new data in applications with measure-valued data; however, computing the Wasserstein barycenter could be extremely computationally intensive. The LOT barycenter, on the other hand, is relatively fast to compute.  In particular, notice the significant difference of computational time between the two methods, as shown in \Cref{tab:barycenter_time}, when we we apply our methods to the tooth dataset.

\begin{table}[h!]
\centering
\begin{tabular}{|l|c|}
\hline
\textbf{Approach} & \textbf{Elapsed Time (seconds)} \\
\hline
True Barycenter & 408835.05 \\
LOT Approximated Barycenter & 187.70\\
\hline
\end{tabular}
\vspace{0.3cm}
\caption{Computational time comparison for barycenter generation approaches}
\label{tab:barycenter_time}
\end{table}

\subsubsection{Functions in the LOT Barycenter Module}

We allow for three main methods that generate barycenters: (1) within classes, (2) between classes, and (3) general barycenters from given weights and measure-valued data.  We go through these methods individually.

The function \texttt{generate\_barycenters\_within\_class} computes barycenters of point clouds belonging to the same class. Given a dataset of LOT embeddings $\mathbf{T} \in \mathbb{R}^{N \times D}$ and corresponding labels $\mathbf{y} \in {1, \dots, C}^N$, the method generates barycenters for each class.  The general methodology for this function first selects LOT embeddings $T_c$, where $c \in 1, \dotsc, C$ corresponds to the classes.  The next step is to determine weights and has two scenarios.  If uniform weights for the barycenter are desired, we set $\lambda_i = \frac{1}{N_c}$ where $N_c$ is the number of LOT embeddings in class $c$ and $i$ corresponds to the $i$th LOT embedding in class $c$. If random weights are desired, generate $\boldsymbol{\lambda} = [\lambda_1, \dots, \lambda_{N_c}]$ such that $\lambda_i \geq 0$ and $\sum_{i=1}^{N_c} \lambda_i = 1$.  Finally, we compute the barycenters for class $c$  by
\begin{equation} 
    \mathbf{b}^{(c)} = \sum_{i=1}^{N_c} \lambda_i (\mathbf{T}_{c})_{i}, 
\end{equation}
where $(\mathbf{T}_{c})_{i}$ is the $i$-th LOT embedding in class $c$.  This results in a set of barycenters $\{\mathbf{b}^{(c)}\}_{c=1}^C$, each representing the average shape within a class.  The code associated with computing the barycenters within classes has signature \texttt{LOTBarycenter.generate\_barycenters\_within\_class(pclouds, labels, weights, uniform, n)} with inputs given by
\begin{itemize}
    \item \texttt{pclouds (np.array)}: A 2D array where each row represents a point cloud. Shape: (\texttt{number\_of\_point\_clouds}, \texttt{number\_of\_features}).
    \item \texttt{labels (np.array)}: A 1D array of labels corresponding to the point clouds in \texttt{pclouds}. Shape: (\texttt{number\_of\_point\_clouds},).
    \item \texttt{weights (list of np.array, optional)}: A list of weight vectors for generating barycenters. If \texttt{None}, random weights are generated. Default is \texttt{None}.
    \item \texttt{uniform (bool, optional)}: Whether to generate a single barycenter per class with uniform weights. Default is \texttt{True}.
    \item \texttt{n (int, optional)}: The number of barycenters to generate per class if weights are not provided. Default is 1.
\end{itemize}
The outputs are given by
\begin{itemize}
    \item \texttt{barycenters (np.array)}: A 2D array where each row represents a barycenter generated by combining point clouds within the same class. Shape: (\texttt{number\_of\_barycenters}, \texttt{number\_of\_features}).
    \item \texttt{output\_labels (np.array)}: A 1D array of labels corresponding to the generated barycenters. Shape: (\texttt{number\_of\_barycenters},).
    \item \texttt{weights (list of np.array)}: A list of weight vectors used to generate the barycenters.
\end{itemize}

The method \texttt{generate\_barycenters\_between\_classes} creates barycenters by combining point clouds from different classes. Given pairs of classes $(c_1, c_2)$, the method generates barycenters that blend features from both classes.

For each class pair $(c_1, c_2)$, we randomly select representative LOT embeddings from class $c_1$ and one from class $c_2$, say $\mathbf{T}_{c_1}$ and $\mathbf{T}_{c_2}$ respectively, and generate weights $\boldsymbol{\lambda} = [\lambda_1, \lambda_2]$ such that $\lambda_i \geq 0$ and $\lambda_1 + \lambda_2 = 1$.  Finally, we compute the barycenter
\begin{equation} 
    \mathbf{b}^{(c_1, c_2)} = \lambda_1 \mathbf{T}_{c_1} + \lambda_2 \mathbf{T}_{c_2}. 
\end{equation}
This process yields barycenters that capture the transitional characteristics between classes $c_1$ and $c_2$.  The signature for this method looks like \texttt{LOTBarycenter.generate\_barycenters\_between\_classes(pclouds, labels, class\_pairs, weights)} where the inputs are given by
\begin{itemize}
    \item \texttt{pclouds (np.array)}: A 2D array where each row represents a point cloud. Shape: (\texttt{number\_of\_point\_clouds}, \texttt{number\_of\_features}).
    \item \texttt{labels (np.array)}: A 1D array of labels corresponding to the point clouds in \texttt{pclouds}. Shape: (\texttt{number\_of\_point\_clouds},).
    \item \texttt{class\_pairs (list of tuple)}: A list of tuples where each tuple contains two class labels to combine for generating barycenters.
    \item \texttt{weights (list of np.array, optional)}: A list of weight vectors for combining point clouds from two classes. If \texttt{None}, random weights are generated. Each weight vector must sum to 1.
\end{itemize}
The outputs are
\begin{itemize}
    \item \texttt{barycenters (np.array)}: A 2D array where each row represents a barycenter generated by combining point clouds from the specified class pairs. Shape: (\texttt{number\_of\_barycenters}, \texttt{number\_of\_features}).
    \item \texttt{representatives (list of tuples)}: A list of tuples where each tuple contains two randomly selected representatives from the specified classes.
    \item \texttt{weights\_list (list of np.array)}: A list of the weight vectors used to generate the barycenters.
\end{itemize}

The method \texttt{generate\_barycenters\_general} allows for the computation of barycenters from arbitrary point clouds using user-defined weights.  Given a set of LOT embeddings $\mathbf{T} \in \mathbb{R}^{N \times D}$ and a set of weight vectors ${\boldsymbol{\lambda}^{(k)}}{k=1}^K$, where each $\boldsymbol{\lambda}^{(k)} \in \mathbb{R}^N$ satisfies $\lambda_i^{(k)} \geq 0$ and $\sum{i=1}^N \lambda_i^{(k)} = 1$, the barycenters are computed by
\begin{equation} 
\mathbf{b}^{(k)} = \sum_{i=1}^N \lambda_i^{(k)} \mathbf{T}_i, \quad \text{for } k = 1, \dots, K.
\end{equation}
This method provides flexibility to generate barycenters based on specific criteria or hypotheses.  The signature for this method is \texttt{LOTBarycenter.generate\_barycenters\_general(pclouds, weights)} where the inputs are
\begin{itemize}
    \item \texttt{pclouds (np.array)}: A 2D array where each row represents a point cloud. Shape: (\texttt{number\_of\_point\_clouds}, \texttt{number\_of\_features}).
    \item \texttt{weights (np.array)}: A 2D array where each row is a weight vector used for generating barycenters by combining the point clouds. Shape: (\texttt{number\_of\_barycenters}, \texttt{number\_of\_point\_clouds}).
\end{itemize}
The outputs are
\begin{itemize}
    \item \texttt{barycenters (np.array)}: A 2D array where each row represents a barycenter generated by combining point clouds in \texttt{pclouds} according to the weight vectors in \texttt{weights}. Shape: (\texttt{number\_of\_barycenters}, \texttt{number\_of\_features}).
\end{itemize}

\subsubsection{Barycenter Generation applied to tooth dataset}

\begin{figure}[h!]
    \centering
        \begin{tabular}{ccc}
        \includegraphics[scale = 0.33]{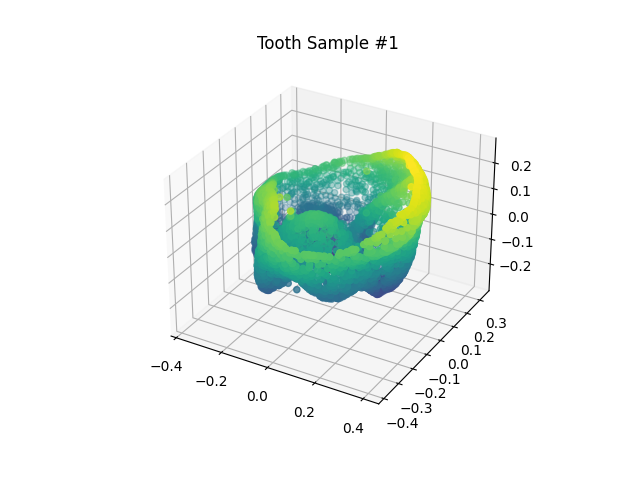} &
        \includegraphics[scale = 0.33]{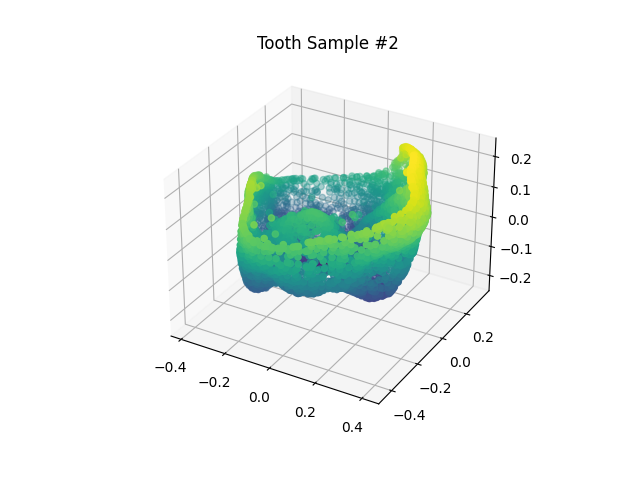} &
        \includegraphics[scale = 0.33]{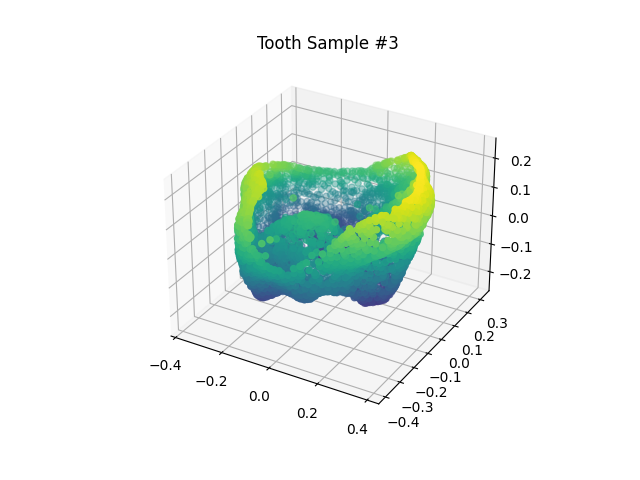} 
        \end{tabular}
    \caption{Fixed Weight Tooth Samples (Tarsius)}
    \label{fig:tooth_samples}
\end{figure}

Given the scarcity and high acquisition cost of ancient teeth data, developing methods to generate new, accurate samples at a relatively low cost is highly beneficial. We explore two approaches for this purpose: the true barycenter and the LOT approximated barycenter.  In particular, we manually input a set of $N = 20$ fixed weights, forming a weight matrix $w_{fixed} \in \mathbb{R}^{3 \times N}$. This matrix is used to generate new samples for each genus using \textbf{only the first three} teeth samples as shown in \Cref{fig:tooth_samples}.

The results were generated by Listing \ref{code:barycenter}.

\begin{lstlisting}[language=Python, caption=Code for barycenter generation., label={code:barycenter}]
# following code works for general situation but for specific case, assume that T_embeddings has only 3 samples per class (the ones from the visual)

# fixed weights for the three teeth
WEIGHTS_FIXED = np.array([
    [1,0,0], [0,1,0], [0,0,1], 
    [0.9, 0.1, 0], [0.9, 0, 0.1], [0.1, 0.9, 0], 
    [0, 0.9, 0.1], [0.1, 0, 0.9], [0, 0.1, 0.9],
    [0.6, 0.2, 0.2], [0.6, 0.3, 0.1], [0.6, 0, 0.4], 
    [0.2, 0.6, 0.2], [0.3, 0.6, 0.1], [0, 0.6, 0.4], 
    [0.2, 0.2, 0.6], [0.3, 0.1, 0.6], [0, 0.4, 0.6], 
    [0.2, 0.4, 0.4], [0.4, 0.2, 0.4], [0.4, 0.4, 0.2]
])

FIXED_SAMPLE_SIZE = 3
WEIGHTS_RAND_LEN = 20

# generate LOT barycenters with fixed weights for teeth in class
G_matrix, G_labels, _ = LOTBarycenter.generate_barycenters_within_class(T_embeddings, T_labels, WEIGHTS_FIXED, uniform=False, n=WEIGHTS_RAND_LEN)

# generate LOT barycenters with random weights for teeth in class
R_matrix, R_labels, WEIGHTS_RAND = LOTBarycenter.generate_barycenters_within_class(T_embeddings, T_labels, None, uniform=True, n=WEIGHTS_RAND_LEN)
\end{lstlisting}

\subsubsubsection{Fixed Weights Generation}

\begin{figure}[h!]
    \centering
        \begin{tabular}{cc}
        \includegraphics[scale = 0.5]{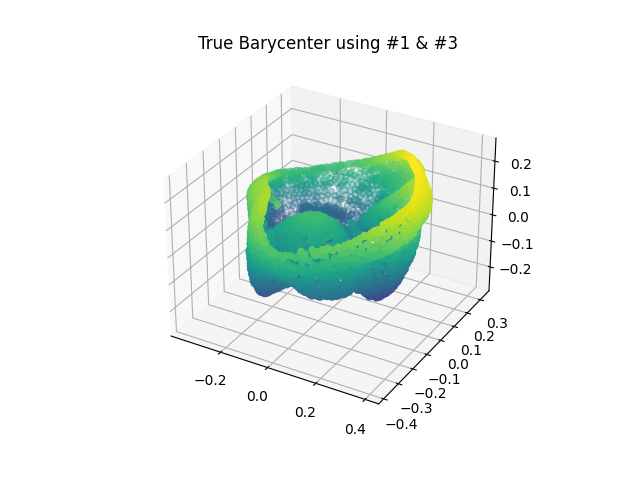} &
        \includegraphics[scale = 0.5]{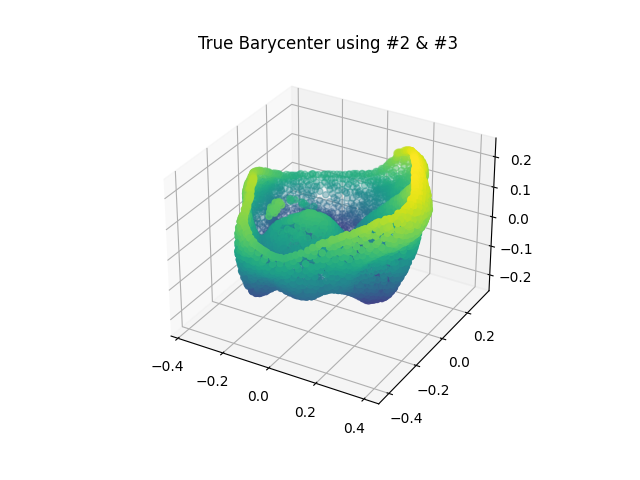} \\
        \includegraphics[scale = 0.5]{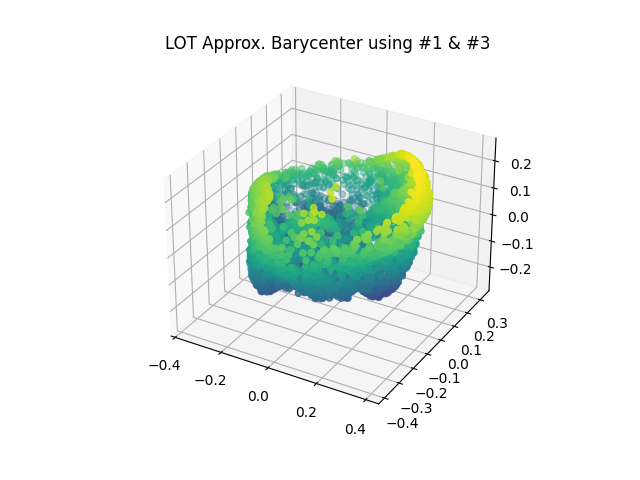} &
        \includegraphics[scale = 0.5]{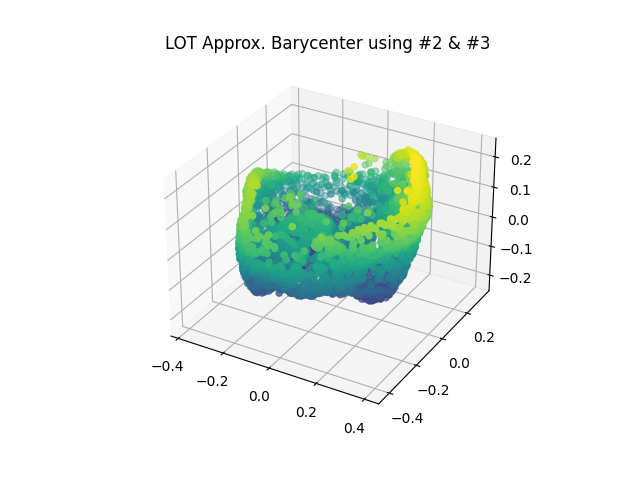}
        \end{tabular}
    \caption{Barycenter Comparision (Fixed Weight)}
    \label{fig:fixed_weight_bary}
\end{figure}

An examination of \Cref{fig:fixed_weight_bary} reveals notable differences between the teeth generated using different sample combinations. Specifically, the Wasserstein barycenter tooth generated from the first and third Tarsius teeth $(\vec w = [0.6, 0, 0.4])$ exhibits distinct characteristics compared to the tooth produced using the second and third Tarsius teeth $(\vec w = [0, 0.6, 0.4])$.  Moreover, note the striking similarity between the LOT-approximated barycenter and the true barycenter, suggesting that the LOT approximation appears to be a reliable method for estimating the barycenter of the sample set, as evidenced by its close correspondence to the true barycenter.

\subsubsubsection{Random Weights Generation}

For each genus $g$, we generate $M = 20$ samples using identical random weights for both approaches. Let $w \in \mathbb{R}^d$ be a weight vector, where each component $w_i \sim U(0, 1)$, i.e., drawn from a uniform distribution over the interval $[0, 1)$, and $d$ corresponds to the number of examples in genus $g$. The same set of uniformly distributed random weights ${w^{(1)}, ..., w^{(M)}}$ is used across both approaches, ensuring a consistent basis for comparison. We then normalize this vector to ensure it sums to one:
\begin{equation}
\hat{w} = \frac{w}{\sum_{i=1}^d w_i}
\end{equation}
Using random weights to generate both Wasserstein barycenters and LOT barycenters, the computational efficiency differs significantly between the two methods, as shown in \Cref{tab:barycenter_time}.

Although the computational efficiency of LOT barycenters is far greater than the Wasserstein barycenters, we must still check how the LOT barycenter compares to the regular Wasserstein barycenter. To do this, we use the relative error rate as a metric.  This metric is calculated through a two-step process. First, we determine the maximum pairwise Wasserstein distance between the LOT-embedded point clouds, which represents the upper bound of potential error in barycenter approximation by considering the diameter of the embedded class. Subsequently, we compute the ratio of the Wasserstein distance between the LOT-approximated barycenter and the true barycenter to this maximum error.  In particular, this relative error is given by
\begin{align*}
    \delta( \{w_k\}_{k=1}^N ) = \frac{ W_2\bigg( \Big( \sum_{k=1}^N w_k \mathbf{T}^(k) \Big)_\sharp \mathbf{x}_r , \arg \min_{\mu} \sum_{k=1}^N w_k W_2^2(\mu, \mathbf{x}_t^{(k)} ) }{ \max_{i,j} W_2 ( \mathbf{T}^{(i)}_\sharp \mathbf{x}_r, \mathbf{T}^{(j)}_\sharp \mathbf{x}_r ) },
\end{align*}
where $\mathbf{x}_r$ corresponds to the current reference measure.  For a particular barycenter problem, this value indicates the percentage of error incurred when using the LOT approximation to compute the barycenter.

Using a single Gaussian reference to generate the LOT embeddings, we use the $M=20$ different random weights to generate $M$ LOT barycenters and compare their relative error with the corresponding true barycenters.  Across the $M$ random weights, we have \Cref{tab:iterlotBary_wassBary_comp} the mean, standard deviation, and sample size of the $M$ experiments, where iteration 0 relates to the case for a single Gaussian reference.

\subsection{Iterative Embeddings}\label{sec:iterative}

Although LOT has great applications, the quality of the LOT embedding depends heavily on the choice of reference measure. To improve the representation of the data, the submodule includes an iterative process to refine the reference measure using LOT barycenters.  To this end, we use the LOT barycenters to estimate the Wasserstein barycenters fast and create good candidates for the reference measures in the LOT embedding.  In the \texttt{LOTEmbedding} submodule, we allow for this functionality with the function \texttt{compute\_barycenter\_embeddings} with the number of iterations specified by the user.  In particular, assume that you have measure-valued data with classes $c \in C$ and $\mathcal{C}_c$ is the set of indices for point clouds in class $c$. For each iteration $j$, the barycenter $\mathbf{b}^{(c)}$ for each class $c$ is computed by minimizing the weighted sum of squared Wasserstein distances
\begin{align*} 
 \frac{1}{\vert \mathcal{C}_c \vert} \sum_{i \in C_c} T^{(i)} = \min_{\mathbf{b}^{(c)}} \quad & \sum_{i \in \mathcal{C}_c} \frac{1}{\vert \mathcal{C}_c \vert } \Vert \mathbf{T}_i - \mathbf{b}^{(c)} \Vert_2.
\end{align*} 
The computed barycenters ${ \mathbf{b}^{(c)}}$ are used as new reference measures for their respective classes. The embeddings are then recomputed using these updated references, enhancing the alignment with the data's intrinsic geometry. In particular, given the target point clouds $\{ \mathbf{x}_t^{(k)} \}_{k=1}^K$, we start with a reference point-cloud $\mathbf{x}_r^{(0)}$ sampled from a multivariate Gaussian that has been centered and scaled using the data.  Now, we iteratively update the reference with the following steps:
\begin{enumerate}
    \item Generate LOT embeddings  $\mathbf{T}^{(k)}$ of all the point-cloud valued data with respect to reference $\mathbf{x}_r^{(j)}$.
    \item Find LOT barycenter of by simply computing $\mathbf{x}_r^{(j+1)} = \Big( \frac{1}{K} \sum_{k=1}^K \mathbf{T}^{(k)} \Big)_\sharp \mathbf{x}_r^{(j)}$.
    \item Set the new reference as $\mathbf{x}_r^{(j+1)}$ and repeat step 1 until you reach the number of iterations.
\end{enumerate}

In the submodule \texttt{LOTEmbedding}, the function used to generate iterative embeddings has function signature \texttt{LOTEmbedding.compute\_barycenter\_embeddings(n\_iterations, embeddings, labels, pclouds, masses, n\_reference\_points, n\_dim)}, where the function has default values as listed.  The inputs of \texttt{LOTEmbedding.compute\_barycenter\_embeddings} are
\begin{itemize}
    \item \texttt{n\_iterations (int)}: The number of iterations for computing barycenter embeddings.
    \item \texttt{embeddings (array or None)}: Initial embeddings of the data points. If \texttt{None}, Gaussian reference measures are generated from the point clouds to initialize the embeddings.
    \item \texttt{labels (array)}: Class labels corresponding to the input data points.
    \item \texttt{pclouds (list of arrays)}: A list of point clouds, where each point cloud is an array of coordinates representing a measure-valued dataset.
    \item \texttt{masses (array)}: Mass distributions corresponding to the point clouds for LOT embedding.
    \item \texttt{n\_reference\_points (int)}: The number of reference points to define the barycenter.
    \item \texttt{n\_dim (int)}: The dimensionality of the space for reshaping reference points.
\end{itemize}
The outputs are given by
\begin{itemize}
    \item \texttt{all\_bary\_embeddings (list of arrays)}: A list where each element contains the embeddings computed at each iteration, including the initial embeddings.
    \item \texttt{all\_emd\_lists (list of lists)}: A list of Earth Mover's Distance (EMD) computations for each iteration. Each inner list corresponds to the embeddings derived from barycenters during the respective iteration.
    \item \texttt{barycenter\_list (list of arrays)}: A list of barycenters computed at each iteration, where each barycenter represents the central point cloud for a given class.
    \item \texttt{barycenter\_label\_list (list of arrays)}: A list of labels corresponding to the barycenters, indicating the class each barycenter belongs to.
\end{itemize}

With code applied to the tooth dataset previously, we can simply run \texttt{LOTEmbedding.compute\_barycenter\_embeddings} to generate the embeddings and corresponding barycenters and run the analysis the same way that was done before.  In particular, this code is run in Listing \ref{code:iterBary}.

\begin{lstlisting}[language=Python, caption=Code for iterative barycenter generation., label={code:iterBary}]
# generate iterative embeddings
all_bary_embeddings, \
    all_emd_lists, \
    barycenter_list, \
    barycenter_label_list = LOTEmbedding.compute_barycenter_embeddings(n_iterations=3, 
                                        embeddings=None,
                                        labels=T_labels, 
                                        pclouds=xt_list, 
                                        masses=None, 
                                        n_reference_points=5000,
                                        n_dim=3
                                        )
\end{lstlisting}

\begin{figure}[h!]
    \centering
        \begin{tabular}{cc}
        \includegraphics[scale = 0.4]{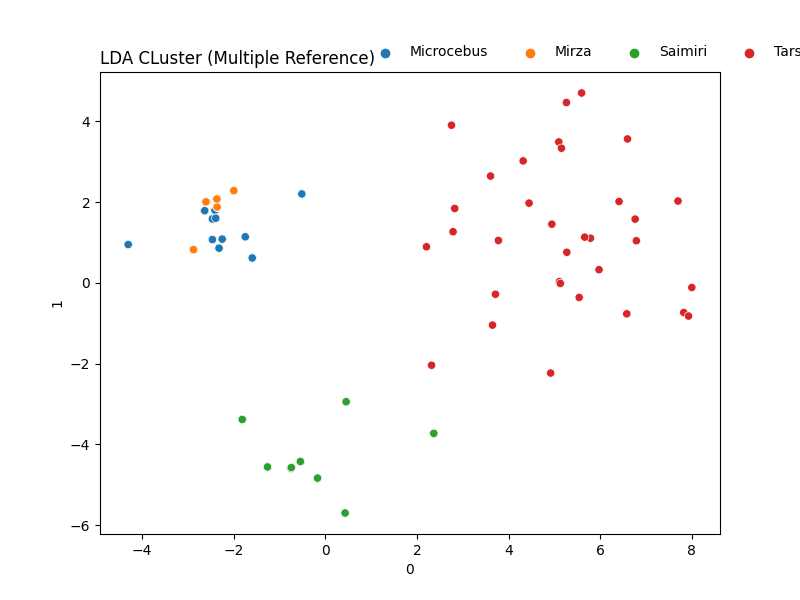} &
        \includegraphics[scale = 0.4]{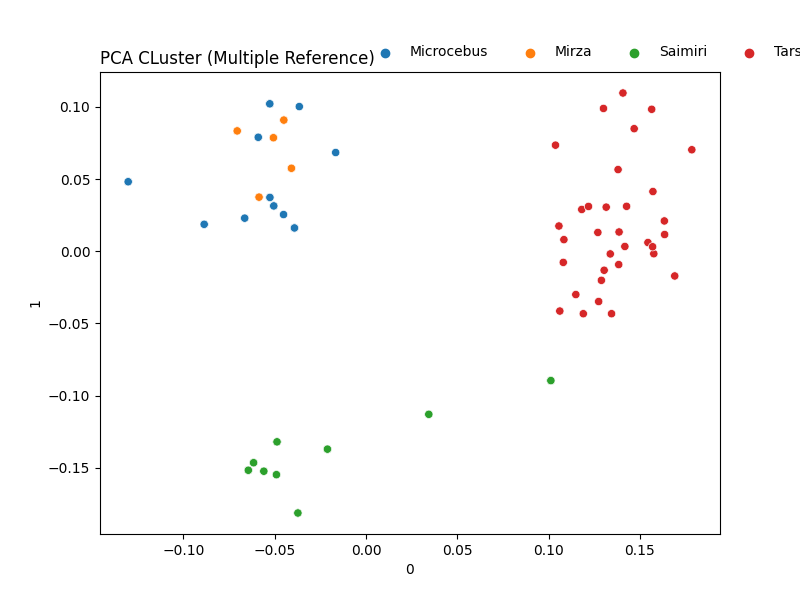}\\
        \includegraphics[scale = 0.35]{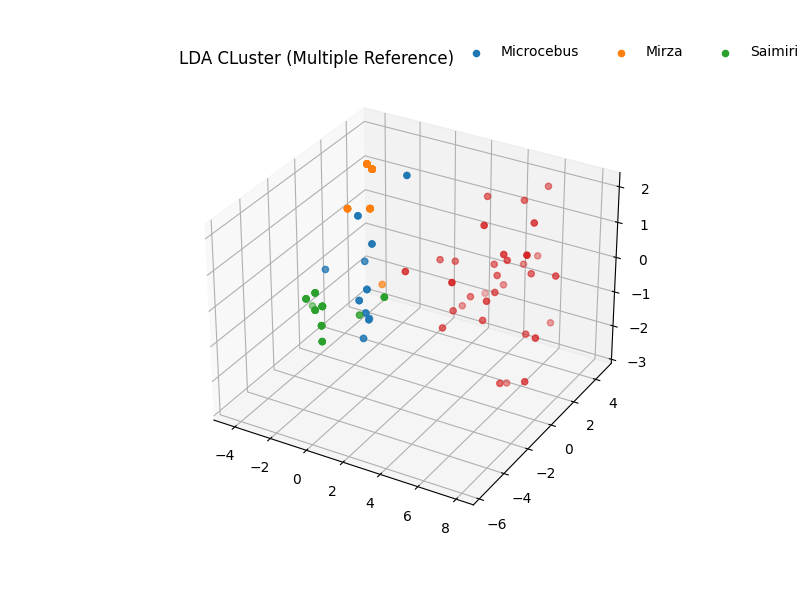} &
        \includegraphics[scale = 0.35]{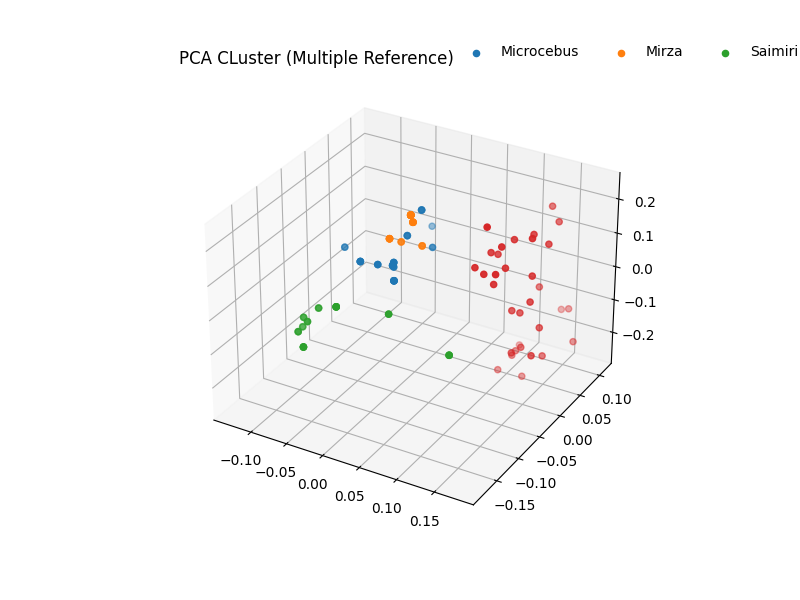}
        \end{tabular}
    \caption{PCA VS. LDA Cluster}
    \label{fig:cluster_imr}
\end{figure}

The downside of generating these iterative embeddings is computational time, which is a trade-off available to the user of pyLOT.  It seems as though the iterated references do slightly better for downstream tasks. In particular, note that the dimensionality reduction in \Cref{fig:cluster_imr} has very slightly better clustering and separation between the Tarsius tooth and other teeth.

Being careful to only generate iterated references from the training data, we repeated the classification problem and saw the cross-validation performance of these models in \Cref{tab:val_performance_iter}.  Similarly, the test performance of these models is summarized in \Cref{tab:test_performance_iter}

\begin{table}[h!]
\centering
\begin{tabular}{|l|c|}
\hline
\textbf{Model} & \textbf{Val. Accuracy} \\
\hline
KNN & 0.981 \\
\textbf{Linear SVM} & \textbf{1.00} \\
RBF SVM & 0.775 \\
\hline
\end{tabular}
\vspace{0.3cm}
\caption{Validation Performance}
\label{tab:val_performance_iter}
\end{table}

\begin{table}[h!]
\centering
\begin{tabular}{|l|c|c|c|c|}
\hline
\textbf{Model} & \textbf{Test Accuracy} & \textbf{Avg. Precision} & \textbf{Avg. Recall} & \textbf{Avg. F1-score} \\
\hline
\textbf{KNN} & \textbf{1.00} & \textbf{1.00} & \textbf{1.00} & \textbf{1.00} \\
Linear SVM & 0.85 & 0.91 & 0.85 & 0.84 \\
RBF SVM & 0.80 & 0.86 & 0.80 & 0.78 \\
\hline
\end{tabular}
\vspace{0.3cm}
\caption{Test Performance}
\label{tab:test_performance_iter}
\end{table}

\begin{figure}[h!]
    \centering
        \begin{tabular}{cc}
        \includegraphics[scale = 0.4]{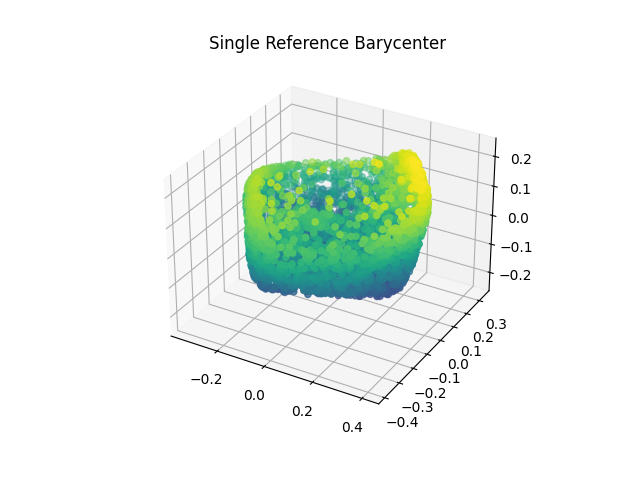} &
        \includegraphics[scale = 0.4]{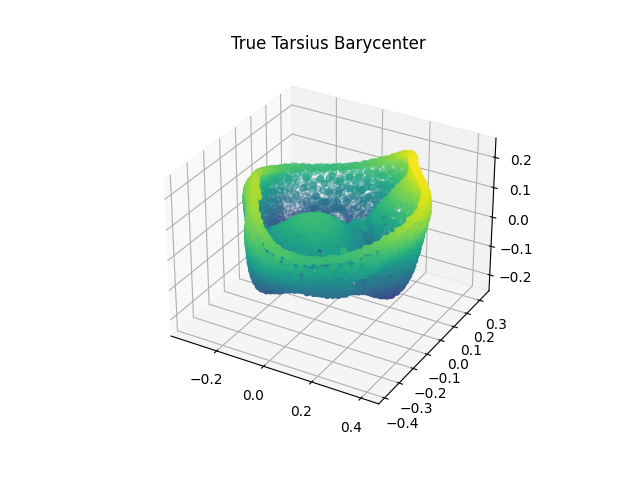} \\
        \includegraphics[scale = 0.4]{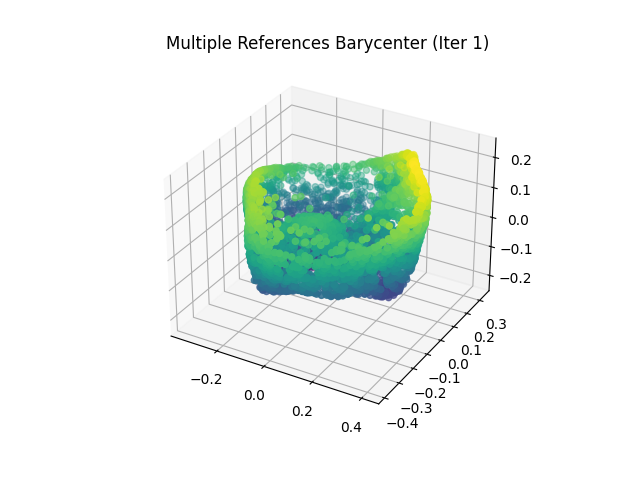} &
        \includegraphics[scale = 0.4]{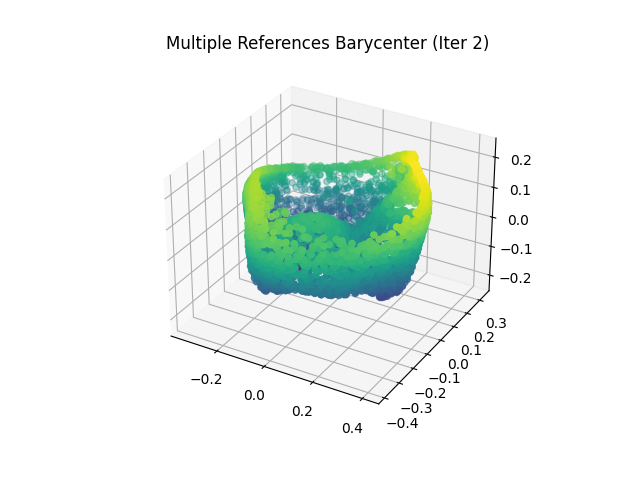}
        \end{tabular}
    \caption{Barycenter Comparison (Iterative)}
    \label{fig:iterative_bary}
\end{figure}

The true usefulness of using iterated references seems to be in generating LOT barycenters as seen in \Cref{fig:iterative_bary}.  Notice that the barycenter seems to improve as we embed with iterated references.  In particular, the sharpness of the LOT barycenter point cloud starts to becomes better aligned with the true Tarsius barycenter.

Similar to the single Gaussian reference case, it is useful to check if the iterative embeddings gain smaller relative error with each iteration.  Including the original single Gaussian reference, we show in \Cref{tab:iterlotBary_wassBary_comp} that indeed smaller relative error is achieved when applying iterative embeddings.  Iteration 0 means that we used a single Gaussian reference measure whilst iteration 1 and 2 denote successive iterative embeddings so that $\mathbf{x}_r$ changes to the LOT barycenter $\mathbf{b}$ computed from the previous iteration. Note that both the sample means and standard deviations of the relative error decrease at iteration 2 quite drastically. This seems to reflect that the barycenter generation in \Cref{fig:iterative_bary} is similar to the other random tests done and holds more generally.

\begin{table}[h!]
\centering
\begin{tabular}{|c|c|c|c|c|c|c|c|}
\hline
\multirow{3}{*}{\textbf{Tooth}} & \multicolumn{3}{c|}{\textbf{Mean}} & \multicolumn{3}{c|}{\textbf{Standard Deviation}} & \multirow{3}{*}{ \textbf{Sample Size}   } \\
\cline{2-7}
& Iteration 0 & Iteration 1 & Iteration 2 & Iteration 0 & Iteration 1 & Iteration 2 & \\
\hline
Microcebus & 0.04722 & 0.04691 & 0.00289 & 0.00308 & 0.00278 & 0.00011 & 11 \\
\hline
Mirza & 0.12196 & 0.15555 & 0.01136 & 0.00912 & 0.00754 & 0.00050 & 5 \\
\hline
Saimiri & 0.03188 & 0.01453 & 0.00251 & 0.00429 & 0.00099 & 0.00016 & 9 \\
\hline
Tarsius & 0.01823 & 0.00882 & 0.00213 & 0.00087 & 0.00035 & 0.00012 & 33 \\
\hline
\end{tabular}
\vspace{0.3cm}
\caption{Statistics of Relative Error Comparing Iterative LOT Barycenter to Wasserstein Barycenter}
\label{tab:iterlotBary_wassBary_comp}
\end{table}

\section{Conclusion}

In this technical report, we introduced pyLOT, a comprehensive Python library designed for efficient Linear Optimal Transport (LOT) embedding of point-cloud data and measure-valued data. We demonstrated its capabilities through extensive experiments on a tooth dataset, highlighting the practical utility and computational efficiency of LOT embeddings in various data analysis tasks.

In particular, we showcase the Python library's capability by showing embeddings of a 3D scanned tooth, applying dimensionality reduction, classification, LOT barycenter approximation, and iterative embeddings.  Through visual inspection, it seemed more optimal to use LOT embeddings generated from the linear program solution to the optimal transport problem rather than Sinkhorn solutions, but this was probably because each of the items in the tooth dataset have sharp features associated to them.  When applying dimensionality reduction on the LOT embedded teeth, we noticed that PCA worked better than LDA. For classification, we demonstrated how the LOT pipeline connects to traditional machine learning methods by applying it to the tooth dataset and saw that the classification does quite well. It seems that the most useful application of this library is by using the LOT barycenter approximation as another method for generative models.  Moreover, the LOT-based method was significantly faster than computing the true Wasserstein barycenter, reducing computation time from several days to a few minutes while maintaining a low relative error.  We were able to generate relatively close samples to the true barycenter of a class of the tooth data by using the LOT barycenter. The iterative embeddings showcased how changing the reference measure to be iterative LOT approximated barycenters enhanced some downstream tasks.  Iterative updates of the reference measure led to a decrease in the relative error between the LOT approximated barycenter and the true barycenter, indicating a closer alignment with the true underlying data distribution.

As a first version of the library, pyLOT offers a powerful set of tools for researchers and practitioners working with point-cloud and measure-valued data. Its ability to efficiently compute LOT embeddings and approximate barycenters opens up new possibilities for data analysis, synthesis, and classification in various scientific fields. Future work will focus on extending the library's namesake of pre-processing measure-valued data to be readily used for black-box methods to new structures of data such as measure-valued time-series.

\bibliographystyle{plain}
\bibliography{references}

\begin{thebibliography}{10}

\bibitem{achlioptas2018learning}
Panos Achlioptas, Olga Diamanti, Ioannis Mitliagkas, and Leonidas Guibas.
\newblock Learning representations and generative models for 3d point clouds, 2018.

\bibitem{aldroubi20}
Akram Aldroubi, Shiying Li, and Gustavo~K. Rohde.
\newblock Partitioning signal classes using transport transforms for data analysis and machine learning.
\newblock {\em Sampl. Theory Signal Process. Data Anal.}, 19(6), 2021.

\bibitem{ambrosio-2013}
Luigi Ambrosio and Nicola Gigli.
\newblock {\em A User's Guide to Optimal Transport}, pages 1--155.
\newblock Springer Berlin Heidelberg, Berlin, Heidelberg, 2013.

\bibitem{arjovsky2017wasserstein}
Martin Arjovsky, Soumith Chintala, and L{\'e}on Bottou.
\newblock {W}asserstein generative adversarial networks.
\newblock In Doina Precup and Yee~Whye Teh, editors, {\em Proceedings of Machine Learning Research}, volume~70, pages 214--223. PMLR, 2017.

\bibitem{Boyer2011}
Doug~M. Boyer, Yonatan Lipman, Elizabeth~M. St~Clair, Jorge Puente, Biren~A. Patel, Thomas~A. Funkhouser, and Jukka Jernvall.
\newblock Algorithms to automatically quantify the geometric similarity of anatomical surfaces.
\newblock {\em Proceedings of the National Academy of Sciences}, 108(45):18221--18226, 2011.

\bibitem{brenier-1991}
Y.~Brenier.
\newblock Polar factorization and monotone rearrangement of vector-valued functions.
\newblock {\em Comm. Pure Appl. Math.}, 44(4):375--417, 1991.

\bibitem{cao2022}
Yueqi Cao and Anthea Monod.
\newblock Approximating persistent homology for large datasets, 2022.

\bibitem{Chen:2017aa}
Yongxin Chen, Filemon~Dela Cruz, Romeil Sandhu, Andrew~L. Kung, Prabhjot Mundi, Joseph~O. Deasy, and Allen Tannenbaum.
\newblock Pediatric sarcoma data forms a unique cluster measured via the earth mover's distance.
\newblock {\em Scientific Reports}, 7(1):7035, 2017.

\bibitem{cloninger2023lotwassmap}
Alexander Cloninger, Keaton Hamm, Varun Khurana, and Caroline Moosmüller.
\newblock Linearized wasserstein dimensionality reduction with approximation guarantees, 2023.

\bibitem{deb2021rates}
Nabarun Deb, Promit Ghosal, and Bodhisattva Sen.
\newblock Rates of estimation of optimal transport maps using plug-in estimators via barycentric projections.
\newblock {\em Advances in Neural Information Processing Systems}, 34:29736--29753, 2021.

\bibitem{Gao2021}
Michael Gao.
\newblock The diffusion geometry of fibre bundles: Horizontal diffusion maps.
\newblock {\em Applied and Computational Harmonic Analysis}, 50:147--215, 2021.

\bibitem{gigli-2011}
Nicola Gigli.
\newblock On {H}{\"o}lder continuity-in-time of the optimal transport map towards measures along a curve.
\newblock {\em Proceedings of the Edinburgh Mathematical Society}, 54(2):401–409, 2011.

\bibitem{khurana2022supervised}
Varun Khurana, Harish Kannan, Alexander Cloninger, and Caroline Moosm{\"u}ller.
\newblock Supervised learning of sheared distributions using linearized optimal transport.
\newblock {\em Sampling Theory, Signal Processing, and Data Analysis}, 21(1), 2023.

\bibitem{kuang2019}
Max Kuang and Esteban~G. Tabak.
\newblock Sample-based optimal transport and barycenter problems.
\newblock {\em Communications on Pure and Applied Mathematics}, 72(8):1581--1630, 2019.

\bibitem{mathews18}
James Mathews, Maryam Pouryahya, Caroline Moosm\"uller, Ioannis~G. Kevrekidis, Joseph~O. Deasy, and Allen Tannenbaum.
\newblock Molecular phenotyping using networks, diffusion, and topology: soft-tissue sarcoma.
\newblock {\em Scientific Reports}, 9, 2019.
\newblock Article number: 13982.

\bibitem{mccann-2001}
R.~J. McCann.
\newblock Polar factorization of maps on {R}iemannian manifolds.
\newblock {\em Geometric {\&} Functional Analysis GAFA}, 11(3):589--608, 2001.

\bibitem{merigot20}
Quentin M\'erigot, Alex Delalande, and Fr\'ed\'eric Chazal.
\newblock Quantitative stability of optimal transport maps and linearization of the 2-{W}asserstein space.
\newblock In Silvia Chiappa and Roberto Calandra, editors, {\em Proceedings of the Twenty Third International Conference on Artificial Intelligence and Statistics}, volume 108 of {\em Proceedings of Machine Learning Research}, pages 3186--3196. PMLR, 26--28 Aug 2020.

\bibitem{moosmueller20}
Caroline Moosm{\"u}ller and Alexander Cloninger.
\newblock Linear {O}ptimal {T}ransport {E}mbedding: {P}rovable {W}asserstein classification for certain rigid transformations and perturbations.
\newblock To appear in: {I}nformation and {I}nference: A {J}ournal of the {IMA}, 2022.

\bibitem{otto2001geometry}
Felix Otto.
\newblock The geometry of dissipative evolution equations: The porus medium equation.
\newblock {\em Communications in Partial Differential Equations}, 26(1-2):101--174, 2001.

\bibitem{park18}
Se~Rim Park, Soheil Kolouri, Shinjini Kundu, and Gustavo~K. Rohde.
\newblock The cumulative distribution transform and linear pattern classification.
\newblock {\em Applied and Computational Harmonic Analysis}, 45(3):616 -- 641, 2018.

\bibitem{PC19}
Gabriel Peyr\'e and Marco Cuturi.
\newblock Computational optimal transport.
\newblock {\em Foundations and Trends in Machine Learning}, 11(5-6):355--607, 2019.

\bibitem{pooladian2021}
Aram-Alexandre Pooladian and Jonathan Niles-Weed.
\newblock Entropic estimation of optimal transport maps.
\newblock {\em arXiv:2109.12004}, 2021.

\bibitem{qi2017}
Charles~R. Qi, Hao Su, Kaichun Mo, and Leonidas~J. Guibas.
\newblock Pointnet: Deep learning on point sets for 3d classification and segmentation, 2017.

\bibitem{qi2017pointnetplus}
Charles~R. Qi, Li~Yi, Hao Su, and Leonidas~J. Guibas.
\newblock Pointnet++: Deep hierarchical feature learning on point sets in a metric space, 2017.

\bibitem{robertson2024}
Sawyer Robertson, Dhruv Kohli, Gal Mishne, and Alexander Cloninger.
\newblock On a generalization of wasserstein distance and the beckmann problem to connection graphs, 2024.

\bibitem{rubner2000earth}
Yossi Rubner, Carlo Tomasi, and Leonidas~J Guibas.
\newblock The earth mover's distance as a metric for image retrieval.
\newblock {\em International journal of computer vision}, 40(2):99--121, 2000.

\bibitem{solomon2014wasserstein}
Justin Solomon, Raif Rustamov, Leonidas Guibas, and Adrian Butscher.
\newblock Wasserstein propagation for semi-supervised learning.
\newblock In {\em International Conference on Machine Learning}, pages 306--314, 2014.

\bibitem{StClair2016}
Elizabeth~M. St~Clair and Doug~M. Boyer.
\newblock Lower molar shape and size in prosimian and platyrrhine primates.
\newblock {\em American Journal of Physical Anthropology}, 161(2):237--258, 2016.

\bibitem{stromme2020barycenter}
Austin~J. Stromme.
\newblock Wasserstein barycenters: statistics and optimiztion.
\newblock S.m. thesis, Massachusetts Institute of Technology, May 2020.
\newblock Cataloged from the official PDF of thesis. Includes bibliographical references (pages 67-71).

\bibitem{villani2008optimal}
C{\'e}dric Villani.
\newblock {\em Optimal Transport: Old and New}, volume 338.
\newblock Springer Science \& Business Media, 2008.

\bibitem{wei13}
Wei Wang, Dejan Slep\v{c}ev, Saurav Basu, John~A. Ozolek, and Gustavo~K. Rohde.
\newblock A linear optimal transportation framework for quantifying and visualizing variations in sets of images.
\newblock {\em Int J Comput Vis}, 101:254--269, 2013.

\bibitem{wang2019}
Yue Wang, Yongbin Sun, Ziwei Liu, Sanjay~E. Sarma, Michael~M. Bronstein, and Justin~M. Solomon.
\newblock Dynamic graph cnn for learning on point clouds, 2019.

\bibitem{werenski2024}
Matthew Werenski, Brendan Mallery, Shuchin Aeron, and James~M. Murphy.
\newblock Linearized wasserstein barycenters: Synthesis, analysis, representational capacity, and applications, 2024.

\bibitem{zaheer2017deepsets}
Manzil Zaheer, Satwik Kottur, Siamak Ravanbakhsh, Barnab{\'a}s P{\'o}czos, Ruslan Salakhutdinov, and Alexander Smola.
\newblock Deep sets.
\newblock In {\em Advances in Neural Information Processing Systems (NeurIPS)}, volume~30, 2017.

\bibitem{zhang2010understanding}
Yin Zhang, Rong Jin, and Zhi-Hua Zhou.
\newblock Understanding bag-of-words model: a statistical framework.
\newblock {\em International Journal of Machine Learning and Cybernetics}, 1(1-4):43--52, 2010.

\end{thebibliography}

\end{document}